\documentclass{ieeeaccess}

\usepackage{subcaption}
\usepackage{csvsimple}
\usepackage{amssymb}
\usepackage[table,dvipsnames]{xcolor}
\usepackage{graphicx}
\usepackage{hyperref}
\usepackage{graphicx}
\usepackage[utf8]{inputenc}

\def\BibTeX{{\rm B\kern-.05em{\sc i\kern-.025em b}\kern-.08em
    T\kern-.1667em\lower.7ex\hbox{E}\kern-.125emX}}
\begin{document}
\history{Date of publication xxxx 00, 0000, date of current version xxxx 00, 0000.}
\doi{10.1109/ACCESS.2017.DOI}

\title{HP-BERT: A framework for longitudinal study of Hinduphobia on social media via language models}

\author{
    Ashutosh Singh\authorrefmark{2,3},
    \and
    Rohitash Chandra\authorrefmark{1,2},~\IEEEmembership{Senior Member, IEEE}
}

\address[1]{Transitional Artificial Intelligence Research Group, School of Mathematics and Statistics, UNSW Sydney, Sydney, Australia} 
\address[2]{Centre for Artificial Intelligence and Innovation, Pingla Institute, Sydney, Australia}
\address[3]{Department of Data Science and Artificial Intelligence, IIIT Naya Raipur, India}

\begin{abstract}
During the COVID-19 pandemic, community tensions intensified, contributing to discriminatory sentiments against various religious groups, including Hindu communities. Recent advances in language models have shown promise for social media analysis with potential for longitudinal studies of social media platforms, such as X (Twitter). We present a computational framework for analyzing anti-Hindu sentiment (Hinduphobia) during the COVID-19 period, introducing an abuse detection and sentiment analysis approach for longitudinal analysis on X. We curate and release a "Hinduphobic COVID-19 XDataset" containing 8,000 annotated and manually verified tweets. We then develop the Hinduphobic BERT (HP-BERT) model using this dataset and achieve 94.72\% accuracy, outperforming baseline Transformer-based language models. The model incorporates multi-label sentiment analysis capabilities through additional fine-tuning. Our analysis encompasses approximately 27.4 million tweets from six countries, including Australia, Brazil, India, Indonesia, Japan, and the United Kingdom. Statistical analysis reveals moderate correlations (r = 0.312-0.428) between COVID-19 case increases and Hinduphobic content volume, highlighting how pandemic-related stress may contribute to discriminatory discourse. This study provides evidence of social media-based religious discrimination during a COVID-19 crisis.
\end{abstract}


\titlepgskip=-15pt

\maketitle


\section{Introduction}\label{Introduction} 

A global pandemic emerged from the severe acute respiratory syndrome coronavirus 2 (SARS-CoV-2) \cite{gorbalenya2020species}, which resulted in the coronavirus disease 2019 (COVID-19)\cite{monteil2020inhibition} with a profound impact on social and economic activities worldwide \cite{das2022socio,kaye2021economic,maital2020global}. One of the most significant aspects of this pandemic has been how social media has influenced public attitudes and behaviours related to the pandemic \cite{jordan2018using}. The COVID-19 pandemic sparked a global public health emergency, leading to a rise in online discussions that exacerbated existing social divisions and sparked new conflicts within communities worldwide \cite{jakovljevic2020covid}. In India's case, a country known for its diverse cultural landscape and inclusive principles, the pandemic exacerbated existing socio-political divisions, particularly concerning communal interactions \cite{bansal2021big}. Although efforts have focused on addressing anti-Islamic sentiments and behaviours \cite{chandra2021virus,bakry2020arguing,ahuja2021labeled,rajan2021insta}, there is another significant but less explored issue: the escalation of Hinduphobia \cite{juluri2020hindu,edwards2017hate} during the COVID-19 crisis. Furthermore, the COVID-19 pandemic has also brought to light the issue of Sinophobia \cite{gao2022sinophobia,zhang2020ignorance,wang2024longitudinal}, {\color{black} but computational analysis of Hinduphobia during global crises remains critically underexplored, representing a significant gap in hate speech detection and social media monitoring systems}.

Hinduism is the world's oldest practising religion \cite{lipner2012hindus} with a population of more than 1.1 billion followers worldwide, mostly residing in India. Hinduism encompasses some of the oldest philosophical (Upanishads and Bhagavad Gita) \cite{gough2013philosophy,grayling2019history}, epic and literary texts (Ramayana and Mahabharata) \cite{kane1966two,winternitz1981history}. Hinduism has contributed to modern science and technology with the invention of the binary and decimal number system \cite{datta1935hi} (also known as Hindu and Arabic numerals) \cite{musa2014origin} and systematic approach to scientific investigation using the philosophy of pramana (evidence) \cite{patil2014methods,oetke2010pramana}. Hinduism is based on non-violent principles that promote vegetarianism \cite{chapple2018animals}, with India having the largest vegetarian population in the world, with the majority of Hindus being partially vegetarian (specific days of vegetarian diet) \cite{chapple2018animals}. Hinduism urges taking a violent stand in self-defence \cite{bodewitz1999hindu} and the fight for dharma (ethics) as promoted in the Bhagavad Gita \cite{kosuta2020ethics}. The term Hinduphobia refers to hate speech and racial and cultural slurs about Hindu religious and cultural practices \cite{juluri2020hindu}. Although limited, a few studies have investigated the nature of Hinduphobia in television and popular culture, which became more prominent with the rise of the Internet and social media \cite{edwards2017hate}.

Hinduism in India withstood its grounds with a history of violence and discrimination, given 800 years of foreign rule, with invasions leading to temple desecration \cite{eaton2000temple}. Thousands of Hindu temples have been destroyed, desecrated, and converted into mosques and churches in history due to foreign rule with religious and political intervention \cite{sears2009fortified}. This has been continuing with Hinduphobia in social media and political discourses, especially in countries where Hindus have been a minority, such as Pakistan and Bangladesh \cite{mahmood2014minoritization,raheja2018neither}. In a majority-Hindu India, anti-Hindu political rhetoric led to the genocide and exodus of Hindus \cite{arora2006living} from the Kashmir valley region in the 1980s \cite{warikoo2023genocide}. Therefore, Hinduphobia is not a recent phenomenon; it has more than a thousand years of history, given the invasions and foreign rule of India \cite{chandra2007history}. Furthermore, Hinduphobic rhetoric has been one of the reasons for the forced conversion of Hindu women in Pakistan \cite{schaflechner2018forced} and gradual changes in ownership and conversion of Hindu temples \cite{schaflechner2018hinglaj}. This has been an issue not just in India but in countries with a Hindu history, such as Indonesia \cite{alfan2016changing}.

The \textit{Kumbh Mela}, which is a major Hindu festival in India, was associated with a high number of COVID-19 cases during the Delta variant outbreak in April 2022 \cite{shukla2021conducting,quadri2021aspect,rocha2021kumbh}. This association, driven by a political media narrative, led to negative portrayals of the Hindu community on social media platforms, including Twitter, Reddit, Instagram, and Facebook. These negative portrayals resulted in trends with hashtags such as \#HinduVirus and \#CoronaHindus, contributing to Hindubhobic statements and hate speech against the Hindu community. Previous studies have examined Hinduphobia in the context of historical events, political dynamics, academic biases, interreligious relations, diaspora experiences, and cultural practices \cite{vemsaniglobal}. Furthermore, there has been an upsurge in hate crimes against Hindus, as demonstrated by the cases of Khaniya Lal, Umesh Kohle, and Nupur Sharma \cite{gitaterrorism}. These incidents highlight the escalating hostility towards Hindus during the pandemic and the need for religious tolerance in society.   However, despite these concerning trends, no comprehensive computational framework has been developed to systematically monitor, detect, and analyze Hinduphobic content at scale during global crises. The COVID-19 pandemic has exacerbated communal tensions in India, which motivates our study to systematically study the rise of Hinduphobic sentiments connected to COVID-19 in India and the rest of the world.

In recent years, LLMs have emerged as powerful tools for analysing and understanding public sentiment on social media. LLMs, such as GPT-3, PaLM, and GPT-4o \cite{brown2020language,chowdhery2022palm} can process vast amounts of textual data, revealing underlying patterns, sentiments, and discourse dynamics that shape public perception and societal narratives \cite{raiaan2024review}. These models can be based on pre-training from a large data corpus and achieve strong performance accuracy in tasks such as decision-making, predictions, text generation, and text summarizations \cite{yao2024survey}. LLMs can detect nuanced language patterns and emerging trends, enabling researchers and policymakers to respond more effectively to rising social and political tensions \cite{krause2020gedi}, taking into account security and privacy \cite{yao2024survey}. Addressing the root causes of discrimination and fostering social cohesion is crucial; however, applying LLMs in this context has a lot of challenges. Data privacy, ethical considerations, and the potential for algorithmic bias must be carefully managed to ensure the deployment of LLMs for societal well-being \cite{hakami2020learning}. LLMs must limit biases and maintain fairness, particularly when dealing with sensitive topics such as communal sentiments and hate speech \cite{breitung2023contextualized}. 

Furthermore, by leveraging the capabilities of LLMs, it is possible to conduct a more granular analysis of sentiment dynamics across different regions, demographics, and periods. Although we can more easily detect sentiments in expressions, the associated meaning is difficult to decipher for models that are not aware of the context and historical development, such as the case of Xenophobic \cite{de2024large,tomasev2024manifestations}, Sinophobic \cite{wang2024longitudinal} and Hinduphobic remarks. This challenge has been demonstrated for code-mixed language settings, such as the case of Hinglish using BERT \cite{nafis2023towards} for hate-speech detection based on a database of Hinglish for misogyny and aggression \cite{bhattacharya2020developing}. Finally, in the case of Hinduphobic remarks, we do not have pre-existing labelled datasets to refine BERT-based models.

 We present a computational framework for detecting and analyzing Hinduphobia during COVID-19. Our study introduces the "Hinduphobic COVID-19 X (Twitter) Dataset" containing 8,000 annotated tweets collected from November 2019 to December 2022, employing rigorous human-in-the-loop annotation methodology \cite{xin2018accelerating}. The dataset categorizes tweets as Hinduphobic (negative) or positive/neutral, with all content related to Hinduism. Details are provided in (Section \ref{Hinduphobic COVID-19 X (Twitter) Dataset}). The dataset is publicly available via Kaggle\footnote{\url{https://www.kaggle.com/datasets/ashutoshsingh22102/hinduphobic-covid-19-x-twitter-dataset-india}}.

We develop the Hinduphobic BERT (HP-BERT) model using multi-stage fine-tuning. First training on our "Hinduphobic COVID-19 X (Twitter) Dataset" for hate detection, achieving 94.72\% accuracy and outperforming five transformer models by 27.51\% to 36.38\% over baseline applications and 5.71\% to 10.49\% over fine-tuned versions. Subsequently, we fine-tune the model on the "SenWave Dataset" \cite{yang2020senwave} to add sentiment analysis capabilities \cite{nogueira2019multi,ma2024fine,liu2021multi}, enabling HP-BERT to both detect Hinduphobic content and analyze sentiment while handling Hinglish content \cite{parshad2016india}. We apply HP-BERT to the "Global COVID-19 Twitter dataset" \cite{lande2024global}, analyzing 27.4 million tweets from six countries: Australia, Brazil, India, Indonesia, Japan, and the United Kingdom. Statistical analysis reveals moderate correlations (r = 0.312-0.428) between COVID-19 case increases and Hinduphobic content volume.  We analyze sentiment polarity scores to examine geographical and temporal variations in Hinduphobic discourse.

The rest of the paper is organised as follows: Section \ref{Related work} reviews the literature relevant to this study. Section \ref{Methods} describes the datasets used, including the data collection and preprocessing processes, along with details of the training and testing of our BERT (HP-BERT) model. Section \ref{Framework} introduces the framework for our approach, and Section \ref{Results} presents the results. Finally, Section \ref{Discussion} provides a comprehensive discussion, and Section \ref{Conclusion} concludes the paper with future research directions.


\section{Related Work}\label{Related work}

\subsection{Sentiment and Semantic Analysis}

Sentiment and semantic analysis have become crucial components of NLP applications, particularly for analysing vast amounts of unstructured textual data from social media platforms \cite{xu2022systematic,rout2018model}. These methods involve understanding the emotional tone (sentiment) and the meaning (semantics) of textual content to extract insights regarding public opinion, trends, and biases \cite{bayrakdar2020semantic}. LLMs have significantly improved the accuracy and scalability of these tasks by leveraging massive datasets and deep model architectures. LLMs can capture complex linguistic patterns and contextual nuances previously challenging to NLP. One of the earliest LLMs, the Bidirectional Encoder Representations from Transformers (BERT) model \cite{devlin2018bert}, has been widely adopted for sentiment analysis tasks \cite{deepa2021bidirectional}. Several customized BERT models have been developed, including news-recommendator BERT \cite{zhang2021unbert}, fake news BERT \cite{kaliyar2021fakebert}, legal-BERT \cite{shao2020bert,chalkidis2020legal}, and medical-BERT \cite{rasmy2021med} for respective domains. The multilingual-BERT (mBERT) \cite{artetxe2019massively} has been pre-trained on monolingual corpora in 104 languages, which has been designed to detect hate speech across multiple languages. DistilBERT \cite{sanh2019distilbert} is a variant of BERT optimized for resource-limited applications. Moreover, advanced variants such as RoBERTa \cite{delobelle2020robbert} have proven effective for nuanced hate speech detection across diverse settings. In the case of the detection of abuse and violence in NLP applications, BERT has been designed for various contexts and categories of hate speech. HateBERT \cite{caselli2020hatebert} is trained on RAL-E, a large-scale dataset of Reddit comments from communities banned for being offensive, abusive, and hateful. These examples of customized BERT models and their variants demonstrate the adaptability and power of LLMs to address the complexities of hate speech detection in specific contexts, showcasing their impact in real-world NLP applications.

Sentiment and semantic analysis have been used to study domains using LLMs, including social media and text analysis. Wang and Chandra \cite{wang2024longitudinal} used BERT-based sentiment analysis and reported that surges in COVID-19 cases correlated with spikes in Sinophobic sentiments. Similarly, Chandra et al. \cite{chandra2024large} used BERT-based sentiment analysis to study trends in The Guardian newspaper coverage of COVID-19, revealing a predominance of negative sentiments. In another study, Chandra and Kulkarni \cite{chandra2022semantic} utilised semantic and sentiment analysis to evaluate the Bhagavad Gita Sanskrit-English translations, comparing verse-by-verse consistency across selected translators. In the field of machine translation, Wang et al. \cite{wang2024evaluation} evaluated Google Translate for its accuracy in translating Mandarin Chinese, applying sentiment and semantic analysis to assess translation precision. Similarly, Shukla et al. \cite{shukla2023evaluation} analysed Sanskrit-to-English translations of the Bhagavad Gita, uncovering inconsistencies in translating philosophical and metaphorical terms by Google Translate. Furthermore, Chandra and Saini \cite{chandra2021biden} modelled the United States 2020 presidential elections using BERT-based sentiment analysis, providing insights into sentiment trends on social media before the elections. Chandra and Krishna \cite{chandra2021covid} compared LSTM and BERT models to track public sentiment during the rise of COVID-19 cases in India. These studies collectively highlight the role of BERT-based models and LLMs in analysing sentiments across social and news media, illustrating their versatility and potential for understanding public opinion.

\subsection{Abuse and Misinformation}

The study of online activity drew significant attention from the research community due to the unique set of challenges faced during the COVID-19 pandemic, with concerns such as anti-vaccination and misinformation \cite{tsao2021social,gabarron2021covid,cascini2022social}. The substantially increased social media activity provided researchers with a wealth of data to analyze public sentiment and discourse. This surge in online interactions also exacerbated stress in society \cite{singh2024optimizing,singh2024machine} as individuals grappled with the rapid spread of misinformation, social isolation, and heightened fears about the pandemic. Chen et al. \cite{chen2020tracking} curated a multilingual coronavirus Twitter dataset with over 123 million tweets for analysing the pandemic's online discourse, reflecting major events and their impacts. Tahmasbi et al. \cite{tahmasbi2021go} conducted one of the first studies on Sinophobic behaviour during the pandemic, analysing over 222 million tweets to reveal the rise in anti-Chinese sentiment.

\begin{figure*}[htbp!]
    \centering
    \includegraphics[width=1\linewidth]{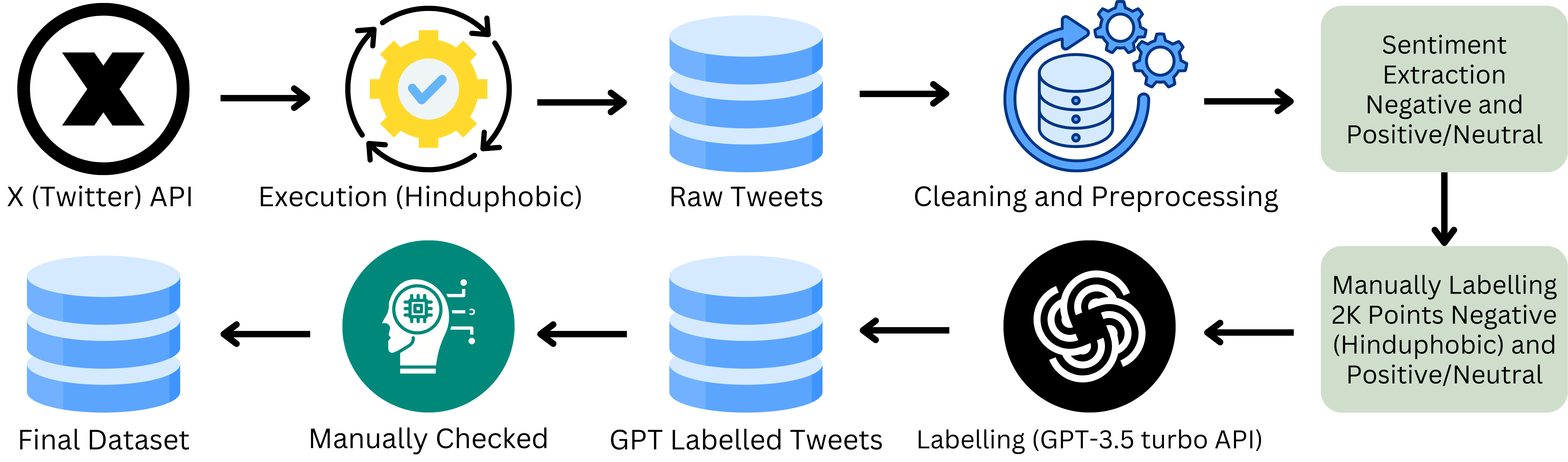}
    \caption{Data collection and processing workflow from initial data extraction via Twitter API to manually labelling data points. This is followed by automatic labelling of the remaining tweets using GPT-3.5 Turbo using a human-in-the-loop strategy \cite{xin2018accelerating}. We removed incorrectly labeled tweets identified during verification to ensure final dataset quality.}
    \label{fig.Hinduphobic_Dataset_Proceess}
\end{figure*}

Cho et al. \cite{cho2020contact} focused on privacy concerns related to COVID-19 contact tracing apps such as TraceTogether, exploring cryptographic protocols and privacy-preserving techniques. The study highlighted the balance between privacy protection and public health benefits, emphasising effective anonymisation and additional privacy measures. Ferrara et al. \cite{ferrara2020covid} investigated the influence of X (Twitter) bots during the early COVID-19 outbreak, particularly in spreading political conspiracies and misinformation. Their curated dataset revealed how bots amplified divisive content during global emergencies. Kouzy et al. \cite{kouzy2020coronavirus} examined COVID-19 misinformation on Twitter, identifying its rapid spread and sources. They found that misinformation often originated from informal groups rather than verified accounts, underscoring the need for robust monitoring and fact-checking mechanisms. Tahmasbi et al. \cite{tahmasbi2021go} analysed Sinophobic content on social media platforms post-COVID-19, using word2vec models to trace the evolution of anti-Chinese sentiment. They highlighted the exacerbation of racial hostility during the pandemic across different online forums.

He et al. \cite{he2021racism} developed a text classifier to detect and mitigate anti-Asian hate speech on Twitter during COVID-19, exploring counter-speech strategies to combat online racism effectively. Chandra et al. \cite{chandra2022deep} employed LSTM models to forecast COVID-19 infections across Indian states, demonstrating their efficacy in predicting cases and providing a two-month-ahead prediction using a recursive prediction strategy. They also highlighted COVID-19's broader impacts, addressing challenges in data reliability and population dynamics. Lande et al. \cite{lande2023deep} applied deep learning techniques for COVID-19 topic modelling for India on Twitter (X) using BERT-based approaches to identify recurring themes such as governance and vaccination. Their study correlated these topics with media coverage across different pandemic phases, providing insights into public discourse and information dissemination during crises.

Despite the extensive research on various aspects of the COVID-19 pandemic, there remains a gap in quantitative analysis regarding Hinduphobic sentiments during this period. However, no computational framework has been developed to systematically analyze Hinduphobia during global crises. This paper aims to fill this gap by examining the prevalence and impact of Hinduphobic discourse on social media platforms amidst the COVID-19 pandemic.

\section{Methods}\label{Methods}

\subsection{Datasets} \label{Datasets}

In this study, we utilise  three datasets that include two existing datasets and one curated by us:
\begin{enumerate}
\item  We curate a new dataset by extracting and processing the \textit{Hinduphobic COVID-19 X (Twitter) dataset} for the classification of Hinduphobic content, with fine-tuning to allow the HP-BERT model to learn the language and sentiment patterns specific to Hinduphobic tweets.
\item  We use the \textit{SenWave dataset} to fine-tune the HP-BERT model for Hinduphobic sentiment analysis. This multi-stage fine-tuning process helped the model adapt to sentiment classification tasks while retaining the domain-specific features learned from the first stage.
\item  We employ the \textit{Global COVID-19 X (Twitter)} dataset for sentiment analysis on Hinduphobic content, enabling the review of sentiment in tweets related to Hinduphobia globally during COVID-19.  
\end{enumerate}

\subsubsection{Hinduphobic COVID-19 X (Twitter) Dataset}  \label{Hinduphobic COVID-19 X (Twitter) Dataset}

\begin{table*}[htbp!]
    \centering
    \small
      \begin{tabular}{|p{2.3cm}|p{14.5cm}|}
        \hline
        \textbf{Keyword} & \textbf{Negative (Hinduphobic) Comments} \\ \hline
        Cow Urine, & You're right. Hindus nearly drink cow urine and bathe in cow dung. I saw videos during COVID suggesting bathing in cow dung as a way to prevent the virus. Disgusting folks.\\ \hline
        Hindu Rituals & People are performing corona puja in Bihar, West Bengal, treating coronavirus as a goddess and offering 9 laddus, 9 cloves, and 9 flowers so the goddess will leave and people won't get sick. No masks, no social distancing. Will superstition ever die? Will scientific temper ever develop? \\ \hline
        Temple & Right after \#AYODHYAVERDICT, the anti-Hindu crowd said, We need world-class hospitals, not temples, but Hindu temples have become resources in this crisis, building COVID hospitals. \#HinduTemplesWithNation \\ \hline
        Kumbh & Why blame the Kumbh just because it's a Hindu festival? Some new research in The Lancet may show that COVID doesn't spread at farmer rallies or religious gatherings of other religions. The judiciary also seems to agree. \\ \hline
        Festivals & BBC News - India Covid: First UK aid arrives as coronavirus deaths mount. Modi and the BJP must take full blame for this tsunami of COVID-19 infections for his irresponsible opening of Hindu festivals that led to this disaster. \\ \hline
          \hline
        \textbf{Keyword} & \textbf{Positive/Neutral (Pro-Hindu) Comments} \\ \hline
        Namaste & For the last few months, we all are fighting against COVID-19. Today, people all over the world are following Hindu practices such as doing Namaste instead of shaking hands, cremating bodies of COVID-19 casualties, etc., to stay safe. \\ \hline
        Food Distribution & It's been 45 days since Hindu groups started distributing homemade food to COVID-affected families. More than 20k meals, never missed a single day. Didn't take a single penny from anyone. Thank you to everyone for being a part of this. Stay safe. \\ \hline
        Yoga & Yoga classes are being conducted for \#COVID19 positive people at Sri Padmavathi Nilayam \#COVID care center, Tirupati. It helps fight against the mental health challenges of people undergoing treatment. \#APFightsCorona \#COVID19Pandemic \\ \hline
        Donation & Hindu temples are doing a great job in these challenging times. Shri Gajanan Maharaj Sansthan, Shegaon, arranged a quarantine facility with 500 beds and daily food for 2,000 people. Money donated and food provided every day! \\ \hline
   Blood Donation & Kali Puja Blood Donation Camp organized by Ahoban Club - Bangur, Bidhannagar, North 24 Parganas. A noble initiative by Sh.~Sajal Bose \& the Puja Committee. More than 100 donors donated blood, used for COVID patients. \\ \hline
    \end{tabular}
 \caption{Keywords and Their Association with Negative (Hinduphobic) and Positive \& Neutral (Pro-Hindu) Sentiments in Comments}
\label{table:negativepositive}
\end{table*}

We extracted Hinduphobic tweets and curated a dataset using the X (Twitter) application programmer interface (API)\footnote{\url{https://developer.x.com/en/docs/x-api}}. The dataset spans from April 2020 to January 2024, capturing the context of the COVID-19 pandemic with tweets featuring both Hinglish (a mix of Hindi and English) and English. Each tweet includes terms such as COVID, COVID-19, or coronavirus, establishing a clear relationship between the pandemic and the Hindu community. (Figure~\ref{fig.Hinduphobic_Dataset_Proceess}) illustrates the complete dataset creation and processing pipeline employed in this study.

We employed targeted keywords and hashtags to identify negative (Hinduphobic) and positive/neutral sentiments associated with Hinduism. We present some of the terms identified and their relevance to Hinduphobic sentiments in (Table~\ref{table:negativepositive}), along with positive/neutral terms. The negative terms such as "Hindu superspreaders" (referring to the Kumbh Mela Hindu festival), "Hindu rituals causing COVID" (related to temple gatherings), and derogatory references such as "piss drinker" (associated with cow urine and dung) were used to malign and blame the Hindu community for the spread of COVID-19. Notable hashtags promoting Hinduphobic content included \#CoronaJihad, \#HindusSpreadCorona, \#HinduCOVIDConspiracy, \#COVID19HinduHate, and \#COVIDHindutva. 

We focused on terms associated with Hindu contributions to COVID-19 relief efforts for the positive or neutral sentiment, such as "Namaste" or "Namaskar" (a Hindu greeting that avoids physical contact) and references to organisations such as ISKCON (International Society of Krishna Consciousness), RSS (Rashtriya Swayamsevak Sangh), Swaminarayan Temple, and Mahavir Temple. These organizations played a significant role in the relief of COVID-19 by providing food, plasma donations, hospital support, and financial aid. Positive hashtags in this category included \#NamasteForSafety, \#HindusForHumanity, \#ISKCONRelief, \#RSSHelpingHands, \#TempleAidCOVID, and \#HinduCOVIDRelief. We derived the full list of keywords from scholarly sources \cite{sudhakar2022anti}, recent computational studies on religious hate speech detection \cite{jahan2023systematic,poletto2021resources,vidgen2020directions}, and validated terminology from peer-reviewed literature on Hinduphobia and religious discrimination \cite{viswanathan2025examining,juluri2015rearming} with explanations for each term's association with negative or positive sentiments available in our supplementary materials\footnote{\url{https://github.com/pinglainstitute/Hinduphobia-COVID-19-beyond}}.

 Our systematic data collection employed stratified sampling across temporal (early pandemic, Delta variant peak, post-vaccine periods), geographic (six countries with significant Hindu populations: Australia, Brazil, India, Indonesia, Japan, and the United Kingdom to ensure comprehensive regional coverage), and content-type dimensions (news, personal opinions, official statements). We collected approximately 150,000 raw tweets using the Academic Research access with consistent 24-hour collection windows and rate limit compliance (300 requests per 15-minute window) to ensure temporal consistency and data quality across the 44-month collection period.

\begin{table*}[htbp!]
    \centering
    \small
    \begin{tabular}{|p{8.5cm}|p{8.5cm}|}
    \hline
    \textbf{Before Preprocessing} & \textbf{After Preprocessing} \\
    \hline
    Rajdeep Sardesai used swastika in his article to portray fascism and nazism to defame Hindus, and Hindus will happily stay silent over such Hinduphobia!! \#Hinduphobia \#Swastika \#Nazism & Rajdeep Sardesai used swastika in his article to portray fascism and nazism to defame hindus and hindus will happily stay silent over such hinduphobia \\
    \hline
    Hindus are the real reason COVID is spreading! Can't believe they still gather at temples! \#HinduSuperspreaders \#COVID19 \#Pandemic & hindus are the real reason covid is spreading, cannot believe they still gather at temples hindusuperspreaders covid19 pandemic \\
    \hline
    Just saw a post about how Hindu rituals are putting everyone at risk! So irresponsible! @username, what do you think? \#HinduRituals \#COVID19 & just saw a post about how hindu rituals are putting everyone at risk so irresponsible [user mention] what do you think hindurituas covid19 \\
    \hline
    LOL, the idea that cow urine cures COVID is just plain ridiculous! \#Hinduphobia \#Superstitions \#COVIDCure & laugh out loud the idea that cow urine cures covid is just plain ridiculous hinduphobia superstitions covidcure \\
    \hline
    OMG, why are they still holding the Kumbh Mela during a pandemic? It's dangerous and selfish! \#KumbhMela \#COVID \#Pandemic & oh my god why are they still holding the kumbh mela during a pandemic it is dangerous and selfish kumbhmela covid pandemic \\
    \hline
    I suspect Congi ecosystem + rabid Hinduphobia of US news media at play. All US papers (WP, NYT) to Right are unanimous in blaming "militant H" for this JNU fracas. \#USMedia \#Hinduphobia & i suspect congi ecosystem rabid hinduphobia of us news media at play all us papers wp nyt to right are unanimous in blaming militant h for this jnu fracas usmedia hinduphobia \\
    \hline
    The thinly veiled Hinduphobia peeping from behind 'secularism' facade is tiring and oh so cliche. \#Secularism \#Hinduphobia & the thinly veiled hinduphobia peeping from behind secularism facade is tiring and oh so cliche secularism hinduphobia \\ 
    \hline
    \end{tabular}
    \caption{Examples of tweets related to Hinduphobic before and after preprocessing.}
    \label{table:processing}
\end{table*}

We conducted a comprehensive bias analysis to address potential sampling bias concerns using chi-square goodness of fit tests comparing our tweet distribution against expected distributions based on Hindu population proportions by country, X user penetration rates, and English language proficiency indices. The analysis revealed significant deviations ($\chi^2 = 47.32, p < 0.001$), primarily due to India's overrepresentation; however, this pattern aligns with the country's large Hindu population (79.8\% of 1.4 billion) and high X usage during COVID-19. Temporal bias controls included consistent 24-hour collection windows across all countries and verification that X API rate limits did not systematically affect data collection during high-volume periods. Our statistical analysis showed no significant correlation between collection date and tweet volume bias (Spearman's $\rho = 0.12, p = 0.34$), indicating consistent sampling across time periods.

After data collection, we preprocessed the dataset to enhance its quality and relevance by removing irrelevant content, spam, and duplicate entries, along with eliminating web links, hashtags, and profile tags \cite{vaswani2017attention, pradha2019effective,symeonidis2018comparative}. We also expanded contractions and abbreviations to ensure consistency in data preparation. Additionally, we transformed emojis into text along with other non-text symbols \cite{duong2021review,mohta2020pre} \cite{kasmaiee2023correcting}. (Table~\ref{table:processing}) presents examples of tweets before and after preprocessing.

We implemented a sequential annotation workflow designed to leverage both human expertise and model capabilities. The process began with expert human annotators manually labeling 2,000 tweets, establishing a high-quality dataset. We developed annotation guidelines through iterative refinement with cultural context experts to ensure consistent interpretation of Hinduphobic content across different cultural and linguistic contexts. Following the manual annotation phase, we fine-tuned GPT-3.5 Turbo using the 2,000 manually annotated tweets as training data.  Our GPT-3.5 Turbo fine-tuning incorporated detailed cultural context, historical background on Hinduphobia, and carefully selected examples representing edge cases and cultural nuances. The training explicitly addressed sarcasm detection, cultural metaphors, and implicit bias recognition to improve automated annotation quality for subsequent labeling tasks.

The fine-tuned GPT-3.5 Turbo model was then deployed to label the remaining tweets from our preprocessed dataset of approximately 15,000 candidates. All GPT-3.5 Turbo-generated labels underwent rigorous human verification, where expert annotators reviewed each prediction and removed incorrectly labeled instances. Through this selective curation process, we retained only 6,000 high-quality GPT-3.5 Turbo labeled tweets that met our annotation standards, ensuring consistency with the manually established gold standard.

Our final annotation quality assessment revealed strong performance metrics: the fine-tuned GPT-3.5 Turbo achieved a 91.7\% agreement rate with manual annotation standards. The Cohen's Kappa between GPT-3.5 labeling and human evaluation was 0.82, indicating substantial agreement.  Additional validation involved cross-annotation of 500 tweets by two independent annotators, achieving Cohen's $\kappa = 0.87$, confirming annotation reliability. This comprehensive quality control process achieved 94.2\% overall annotation quality using Cohen's Kappa-based scoring methodology.

Our final annotation quality assessment revealed strong performance metrics: the fine-tuned GPT-3.5 Turbo achieved a 91.7\% agreement rate with manual annotation standards. The Cohen's Kappa between GPT labeling and human evaluation was 0.82, indicating substantial agreement. Additional validation involved cross-annotation of 500 tweets by two independent annotators, achieving Cohen's $\kappa = 0.87$, confirming annotation reliability. This comprehensive quality control process achieved 94.2\% overall annotation quality using Cohen's Kappa-based scoring methodology. (Table~\ref{tab:annotation_summary}) provides a comprehensive overview of our multi-phase annotation workflow, detailing the progression from initial manual annotation through AI-assisted labeling to final quality validation. 

\begin{table}[h]
\centering
\scriptsize
\caption{Dataset Annotation Summary and Quality Metrics}
\label{tab:annotation_summary}
\begin{tabular}{l|c|l|c|c}
\hline
\textbf{Phase} & \textbf{Tweets} & \textbf{Method} & \textbf{Agreement} & \textbf{Quality} \\
\hline
Manual & 2,000 & Expert Human & - & 100\% \\
\hline
GPT Fine-tuning & 2,000 & Training Data & - & - \\
\hline
GPT Labeling & 15,000 & Fine-tuned Model & 91.7\% & - \\
\hline
Human Verification & 6,000 & Quality Selection & $\kappa$ = 0.82 & 94.2\% \\
\hline
Cross-Validation & 500 & Dual Annotators & $\kappa$ = 0.87 & - \\
\hline
\textbf{Final Dataset} & \textbf{8,000} & \textbf{Hybrid Approach} & \textbf{$\kappa$ = 0.87} & \textbf{94.2\%} \\
\hline
\end{tabular}
\end{table}

\begin{figure*}[htbp!]
    \centering
    \includegraphics[width=1\linewidth]{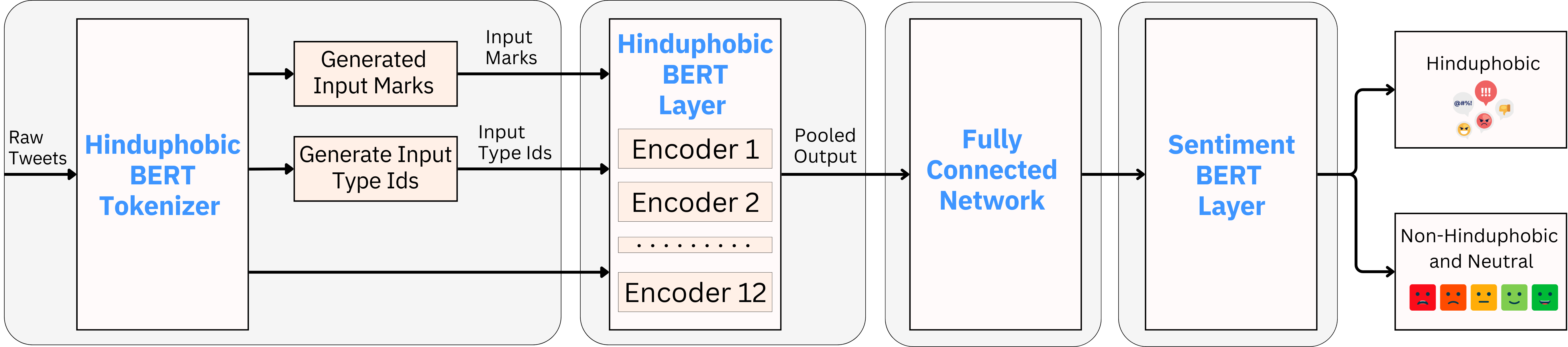}
\caption{HP-BERT model architecture for detecting Hinduphobic sentiment, featuring tokenisation (encoding) of Hinglish/English X  using a set of encoders. It includes a fully connected layer for classifying X as Hinduphobic or Non-Hinduphobic/Neutral, and the final layer provides sentiment classification into different categories.}
    \label{fig.HP_BERT_Layer}
\end{figure*}

The "Hinduphobic COVID-19 X (Twitter) Dataset" is publicly available on Kaggle under a Creative Commons license. To ensure user privacy and platform compliance, the dataset contains only tweet IDs and classification labels, with all personally identifiable information (usernames, profile details, location data) removed. The final dataset contains 8,000 tweets (4,600 labeled as Hinduphobic and 3,400 as positive/neutral), published on Kaggle \footnote{\url{https://www.kaggle.com/datasets/ashutoshsingh22102/hinduphobic-covid-19-x-twitter-dataset-india}}. This rigorous annotation process was necessary due to the challenge of accurately identifying hate speech, which is inherently language and culture-dependent. Existing models, such as HateBERT \cite{caselli2020hatebert} and COVID-HateBERT \cite{li2021covid}, may struggle to detect nuanced expressions of Hinduphobic content with hidden meanings. Correctly identifying instances of hate speech is essential to maintaining data integrity, ensuring accurate interpretation, and minimizing the risk of inaccuracies. This dataset serves as the foundation for fine-tuning the HP-BERT model, specifically designed to improve the detection of Hinduphobic content.

\subsubsection{SenWave dataset}

The SenWave dataset \cite{yang2020senwave} contains over 104 million tweets and Weibo messages related to COVID-19 in six different languages: English, Spanish, French, Arabic, Italian, and Chinese. The data was collected between March 1 and May 15, 2020, to analyse social media posts surrounding the pandemic. It includes 10,000 English and 10,000 Arabic tweets labelled into ten sentiment categories: optimistic, thankful, empathetic, pessimistic, anxious, sad, annoyed, denial, official report, and joking. This comprehensive labelling facilitates in-depth analysis of the diverse emotional responses exhibited during the pandemic. The effectiveness of the SenWave dataset has been validated in two prior studies \cite{chandra2021covid,chandra2023analysis}, demonstrating its efficacy for sentiment analysis tasks. After fine-tuning our BERT model (HP-BERT) with the Hinduphobia dataset for Hinduphobic content classification, we will apply multi-stage fine-tuning using the SenWave dataset. Fine-tuning will enable the model to perform sentiment analysis on the Global COVID-19 X (Twitter) dataset, refining its understanding of sentiment in the context of Hinduphobic content during the pandemic.

\subsubsection{Global COVID-19 X (Twitter) Dataset}

The Global COVID-19 X (Twitter) dataset \cite{lande2024global} is a comprehensive collection of tweets related to the COVID-19 pandemic, spanning from March 2020 to February 2022. This dataset includes tweets from six countries: Australia, Brazil, India, Indonesia, Japan, and the United Kingdom, as outlined in (Table \ref{tab:tweetcounts}). It captures various public opinions and sentiments about the pandemic, enabling detailed analysis across diverse cultural and regional contexts. This dataset has also been utilised by Chen et al. \cite{wang2024longitudinal} for sentiment analysis related to Sinophobia (anti-China sentiments) and Chandra et al. \cite{lande2023deep} for the analysis of antivaccine-related tweets during COVID-19 across different countries. We processed this data set similarly to the Hinduphobic data set, ensuring standardized text input essential for the model to accurately capture contextual nuances in the data. The processed dataset will be used to analyze Hinduphobic tweets across different countries, as well as sentiment analysis in various regions using our  HP-BERT model.

 \begin{table}[htbp!]
    \centering
    \begin{tabular}{c|c}
    \hline
    \textbf{Country Name} & \textbf{Number of Tweets} \\
    \hline
    Australia & 3,212,464 \\      \hline
    Brazil & 943,913 \\      \hline
    India & 5,411,294 \\      \hline
    Indonesia & 229,935 \\      \hline
    Japan & 644,510 \\      \hline
    United Kingdom & 16,958,866 \\      
    \hline
    \end{tabular}
    \caption{Tweet counts from six different countries.}
    \label{tab:tweetcounts}
\end{table}

\begin{figure*}[htbp!]
    \centering
    \includegraphics[width=1\linewidth]{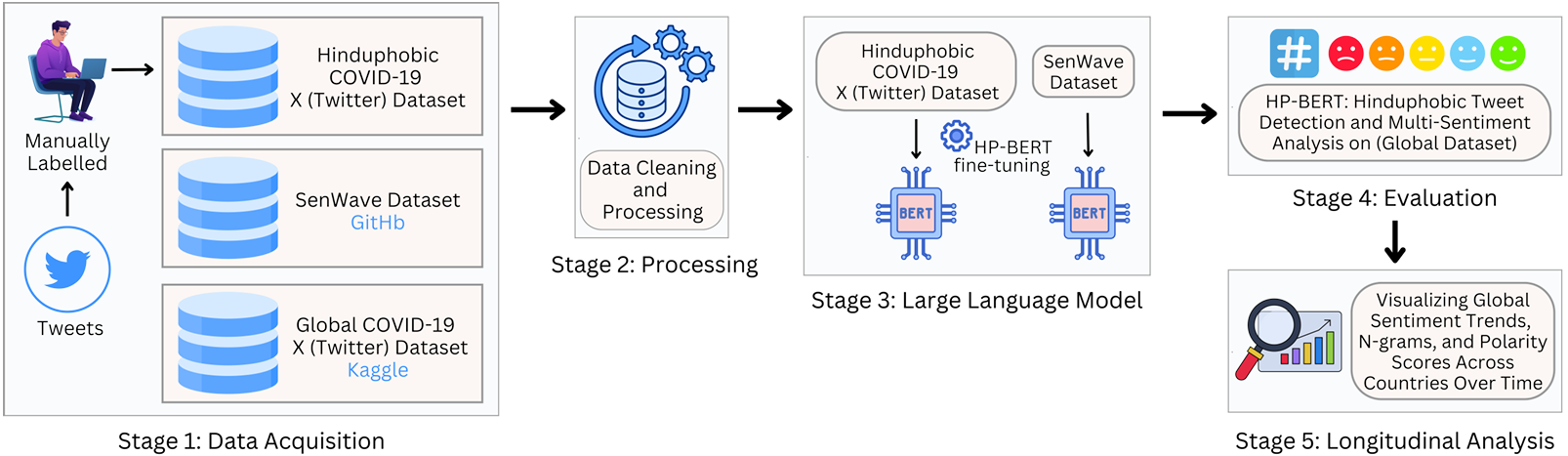}
    \caption{A multi-stage process for HP-BERT model training and analysis: Stage 1 involves dataset collection, Stage 2 preprocesses data, Stage 3 fine-tunes HP-BERT for binary and sentiment classification, Stage 4 applies classification and trend analysis, and Stage 5 conducts sentiment analysis with visualisation of trends and polarity.}
    \label{fig:framework}
\end{figure*}

\subsection{Hinduphobic-BERT Model}

BERT \cite{devlin2018bert} is an advanced language model designed to understand the contextual relationships between words by simultaneously considering both left and right contexts in a sentence. BERT uses a bidirectional attention mechanism based on the Transformer architecture \cite{mozafari2020bert}, allowing it to read an entire sentence at once and understand the context more effectively. The attention mechanism enables BERT to weigh the importance of different input parts, making it superior for language understanding tasks such as sentiment analysis \cite{alaparthi2021bert}. The role of BERT in this context is to generate contextualised embeddings for each tweet in the dataset. The pre-training on a massive corpus enables BERT to capture nuanced language features, making it highly suitable for sentiment analysis tasks \cite{lande2023deep},  and detecting hate speech \cite{saleh2023detection} and bias patterns. Furthermore, the BERT model can be easily extended to domain-specific tasks by fine-tuning with specialised datasets.

We develop a Hinduphobic BERT (HP-BERT) model to classify tweets containing Hinduphobic sentiments. The HP-BERT model architecture for detecting Hinduphobic sentiment is shown in (Figure \ref{fig.HP_BERT_Layer}). The process begins with the Hinduphobic BERT Tokenizer, which tokenises Hinglish (a blend of Hindi and English) and English tweets into input identifiers (IDs), input type IDs, and input mask. Input IDs represent the tokenised text in numeric form, enabling the model to process it, while input type IDs distinguish between different parts of the input, such as multiple sentences. The attention mask ensures that the model focuses only on meaningful tokens, ignoring padding. These inputs are then passed through a 12-layer encoder to generate contextualized embeddings that are later fed into the fully connected layer to classify each tweet as either Hinduphobic or Non-Hinduphobic/Neutral. We then use an additional layer for sentiment classification by training (refining) using the SenWave dataset \cite{yang2020senwave}. The HP-BERT model enables a comprehensive sentiment analysis of tweets by combining Hinduphobic detection with sentiment classification, which also includes the computation of a sentiment polarity score based on predefined weights in (Table \ref{tab:sentiment_weights}). This approach allows the model to detect Hinduphobic content while providing detailed sentiment categorization and sentiment polarity analysis.

\begin{table}[htbp!]
    \centering
 \begin{tabular}{p{0.2\textwidth}p{0.1\textwidth}}
        \hline
        \textbf{Sentiment} & \textbf{Weight} \\
        \hline
        Optimistic & 2 \\      \hline
        Thankful & 3 \\      \hline
        Empathetic & 0 \\      \hline
        Pessimistic & -4 \\      \hline
        Anxious & -2 \\      \hline
        Sad & -3 \\      \hline
        Annoyed & -1 \\      \hline
        Denial & -5 \\      \hline
        Official & 0 \\      \hline
        Joking & 1 \\     
        \hline
    \end{tabular}
    \caption{Weight ratios for different sentiments for polarity score calculation.}
    \label{tab:sentiment_weights}
\end{table}

\begin{table*}[htbp!]
\centering
\resizebox{\textwidth}{!}{ 
\begin{tabular}{l|l|c|c|c|c}
\hline
\textbf{Data} & \textbf{Method} & \textbf{Accuracy (Mean $\pm$ Std)} & \textbf{F1 (Mean $\pm$ Std)} & \textbf{Precision (Mean $\pm$ Std)} & \textbf{Recall (Mean $\pm$ Std)} \\ \hline
Hinduphobic Dataset (20/80) & BERT & 0.9053 $\pm$ 0.0051 & 0.9048 $\pm$ 0.0053 & 0.9059 $\pm$ 0.0050 & 0.9050 $\pm$ 0.0054 \\ \hline
Hinduphobic Dataset (40/60) & BERT & 0.9215 $\pm$ 0.0042 & 0.9210 $\pm$ 0.0044 & 0.9221 $\pm$ 0.0041 & 0.9213 $\pm$ 0.0045 \\ \hline
Hinduphobic Dataset (60/40) & BERT & 0.9426 $\pm$ 0.0035 & 0.9421 $\pm$ 0.0037 & 0.9430 $\pm$ 0.0034 & 0.9423 $\pm$ 0.0038 \\ \hline
Hinduphobic Dataset (80/20) & BERT & 0.9472 $\pm$ 0.0028 & 0.9465 $\pm$ 0.0030 & 0.9474 $\pm$ 0.0029 & 0.9467 $\pm$ 0.0031 \\ \hline
\end{tabular}}
\caption{Evaluation metrics (mean and standard deviation) for HP-BERT across 30 experimental runs. Metrics include accuracy, F1 score, precision, and recall for various training/testing splits on Hinduphobic COVID-19 X (Twitter) Dataset.}
\label{tab:HPBERT_evaluation_metrics}
\end{table*}

\section{Framework} \label{Framework}

Our Hinduphobic tweet detection and sentiment analysis framework involves multiple stages and components, as illustrated in (Figure \ref{fig:framework}). 

In Stage 1, we obtain three datasets to fine-tune and evaluate our HP-BERT model for sentiment classification, including the i. Hinduphobic COVID-19 X (Twitter) Dataset \cite{ashutosh_singh_rohitash_chandra_2024}, ii. SenWave Dataset \cite{yang2020senwave}, and iii. Global COVID-19 X (Twitter) Dataset \cite{lande2024global}. In the case of the "Hinduphobic COVID-19 X (Twitter) Dataset", we use HITL for manual labelling of  8000 tweets and also develop a model for classification of the leftover tweets. 

In Stage 2, we preprocess the three datasets by removing irrelevant content, spam, and duplicate entries, as well as eliminating web links, hashtags, and profile tags. We also expanded contractions and abbreviations to ensure consistency across the data. Additionally, we convert the emojis to text along with other symbols for expressions used in social media. 

In Stage 3, we first fine-tune a pre-trained BERT model using the "Hinduphobic COVID-19 X (Twitter) Dataset" to detect and classify Hinduphobic content in tweets. Afterwards, we further fine-tune the model on the "SenWave dataset" to incorporate sentiment classification with an additional layer, as shown in (Figure \ref{fig.HP_BERT_Layer}). Another layer is added for sentiment classification, allowing the model to extract sentiments associated with the tweet to enable the HP-BERT model to perform both Hinduphobic content detection and sentiment analysis (multi-label sentiment classification).

In Stage 4, we utilise the trained HP-BERT model for "the Global COVID-19 X (Twitter) Dataset" to examine the volume of Hinduphobic content by examining various dimensions: total counts, country-wise distribution, and month-wise trends. Additionally, we conduct bigraph, bigram, and trigram analyses both globally and individually for each country. This multi-faceted analysis helps us understand how Hinduphobic content evolved and varied geographically. The insights gained from these analyses provide a valuable understanding of global patterns and trends in Hinduphobic discourse during the pandemic. Additionally, we apply HateBERT \cite{caselli2020hatebert} to assess its effectiveness in detecting hate speech and abusive content when related to Hinduphobia. 

In Stage 5, we apply our HP-BERT model to perform sentiment analysis on Hinduphobic content, which involves conducting a longitudinal analysis and generating various plots to examine the sentiment distribution and trends across different regions, including sentiment polarity scores. Polarity scores are calculated using the TextBlob library, and custom weight ratios for different sentiment labels, as shown in (Table \ref{tab:sentiment_weights}) are applied to enhance the analysis. This stage provides deeper insights into how sentiments in Hinduphobic content fluctuate over time and vary geographically.

\section{Results} \label{Results}

\subsection{Technical setup}

We use a dropout regularization rate of \(d = 0.2\) to prevent overfitting in BERT, obtained from Huggingface\footnote{\url{https://huggingface.co/docs/transformers/en/model_doc/bert}}. The model is trained for 10 epochs with a learning rate of \(2 \times 10^{-5}\) and a batch size of 8. Additionally, weight decay regularization with a rate of \(\gamma = 0.01\) is applied to further mitigate overfitting. We utilize Python-based libraries, including Transformers, NumPy, Pandas, Matplotlib, and PyTorch. The training process is accelerated using an NVIDIA GeForce RTX 2080 Ti GPU (graphics processing unit)with 11264 MiB of memory and an AMD Ryzen 7 5800H processor, along with 16 GB of system memory.

\subsubsection{Fine-tuning the HP-BERT Model}

We fine-tuned the HP-BERT model for Hinduphobia detection and sentiment analysis using the "Hinduphobic COVID-19 X (Twitter) Dataset", which contains 8,000 labeled tweets. The dataset was split into four different training/testing configurations (randomly). The fine-tuned model was evaluated on the test set to assess its performance in classifying Hinduphobic tweets. (Table \ref{tab:HPBERT_evaluation_metrics}) presents the results from 30 independent experimental runs, reporting the mean and standard deviation for accuracy, F1 score, precision, and recall. The results demonstrate strong performance, with high mean accuracy and balanced F1 scores, indicating the model's ability to maintain precision and recall. Next, we used the fine-tuned HP-BERT model to predict Hinduphobic tweets in a global dataset.

In (Table \ref{tab:key_evaluation_metrics}), we present the results of the fine-tuned HP-BERT model on the SenWave dataset for sentiment analysis, offering additional analysis following Hinduphobic tweet detection. Since this is a multi-label sentiment analysis task, conventional metrics are not applicable. Instead, we report the Label Ranking Average Precision (LRAP) score, F1-score, Jaccard score, and Hamming loss. These metrics have been widely used in earlier studies \cite{lande2023deep,chandra2021covid}, and our results further validate their findings.

\begin{table}[htbp!]
\centering
\begin{tabular}{l|c}
\hline
\textbf{Metric} & \textbf{Value} \\ \hline
Hamming Loss & 0.1476 \\ \hline
Jaccard Score & 0.5013 \\ \hline
Label Ranking Average Precision (LRAP) & 0.7501 \\ \hline
F1 Score (Macro) & 0.5187 \\ \hline
F1 Score (Micro) & 0.5834 \\ \hline
\end{tabular}
\caption{Evaluation metrics of the HP-BERT model after fine-tuning on the SenWave dataset for sentiment analysis.}
\label{tab:key_evaluation_metrics}
\end{table}

\subsection{Comparative Model Analysis}

 We conducted a comprehensive evaluation against recent state-of-the-art Transformer-based models to validate the effectiveness of our HP-BERT model. We stress that HP-BERT incorporates multi-task capabilities (Hinduphobia detection and sentiment analysis), while baseline models focus on binary task classification. We designed our evaluation to ensure fair comparison while highlighting our model's enhanced functionality. We conducted two distinct evaluation scenarios: (1) Baseline Model Comparison, where we directly downloaded pre-trained models from HuggingFace Hub and evaluated them on our "Hinduphobic COVID-19 X (Twitter) Dataset" without any domain-specific adaptation, and (2) Fine-tuned Model Comparison, where we fine-tuned all baseline models using our "Hinduphobic COVID-19 X (Twitter) Dataset" under identical conditions and hardware configurations used for the HP-BERT model.

We selected five baseline models representing diverse architectural innovations and training paradigms: RoBERTa \cite{liu2019roberta} with optimized BERT training procedures and dynamic masking; DeBERTa \cite{he2020deberta} featuring disentangled attention mechanisms that separately encode content and position information; ELECTRA \cite{clark2020electra} employing replaced token detection objectives for efficient pre-training; DistilBERT \cite{sanh2019distilbert} demonstrating knowledge distillation approaches with a parameter reduction of 60\% while maintaining competitive performance; and HateBERT \cite{caselli2020hatebert}, a domain-specific model pre-trained on offensive content for the detection of hate speech.

 The baseline models were evaluated using their default configurations from the HuggingFace Hub without any domain-specific adaptation. We accessed the models through the \texttt{transformers.pipeline()} API, employing sentiment analysis as a proxy for hate speech detection. All baseline models were subsequently fine-tuned on the "Hinduphobic COVID-19 X (Twitter) Dataset" for the binary Hinduphobia detection task, using identical hyperparameters and training configurations as those employed for HP-BERT training. We accessed performance using the classification accuracy, precision, recall, and F1-score.  

 (Table~\ref{tab:model_comparison_outofbox}) demonstrates substantial performance disparities between HP-BERT and baseline model configurations. HP-BERT achieved 94.72\% accuracy, significantly outperforming all baselines: HateBERT (67.21\%), RoBERTa (66.45\%), DeBERTa (65.18\%), ELECTRA (63.92\%), and DistilBERT (58.34\%). The 27.51\% performance gap over HateBERT, despite its hate speech specialization, highlights the critical importance of cultural and linguistic specificity in hate speech detection. These results validate our hypothesis that general-purpose models, even those trained for hate speech detection, do not perform well on culturally-specific tasks such as Hinduphobia detection. 

\begin{table}[h]
\centering
\begin{tabular}{lcccc}
       \hline
Model & Accuracy & Precision & Recall & F1-Score \\
       \hline
DistilBERT \cite{sanh2019distilbert} & 0.5834 & 0.5782 & 0.5886 & 0.5815 \\      \hline
ELECTRA \cite{clark2020electra} & 0.6392 & 0.6347 & 0.6436 & 0.6375 \\      \hline
DeBERTa \cite{he2020deberta} & 0.6518 & 0.6483 & 0.6552 & 0.6502 \\
      \hline
RoBERTa \cite{liu2019roberta} & 0.6645 & 0.6591 & 0.6698 & 0.6618 \\
      \hline
HateBERT \cite{caselli2020hatebert} & 0.6721 & 0.6507 & 0.6693 & 0.6802 \\
      \hline
HP-BERT (Ours) & 0.9472 & 0.9474 & 0.9467 & 0.9465 \\
       \hline
\end{tabular}
\caption{Baseline model comparison on Hinduphobic tweets detection}
\label{tab:model_comparison_outofbox}
\end{table}

When all models underwent identical fine-tuning on our dataset, the results shown in (Table~\ref{tab:model_comparison_finetuned}) indicate that performance gaps reduced significantly but remained substantial, providing insights into both domain adaptation importance and architectural advantages. HP-BERT maintained superior performance (94.72\% accuracy), followed by HateBERT (89.01\%), RoBERTa (88.12\%), DeBERTa (87.45\%), ELECTRA (86.98\%), and DistilBERT (84.23\%). The convergence of baseline models to the 84-89\% range demonstrates that domain-specific fine-tuning significantly improves performance across all architectures, while HP-BERT's sustained advantage (5.71\% over HateBERT, 6.60\% over RoBERTa) indicates that our multi-stage fine-tuning approach and cultural context modeling provide architectural benefits beyond standard domain adaptation.

HP-BERT incorporates sentiment analysis capabilities through its multi-task learning approach, providing a significant advantage beyond classification performance. The multi-stage fine-tuning enables additional sentiment analysis capabilities without compromising primary task performance. This dual functionality allows comprehensive discourse analysis, as demonstrated in our longitudinal study of global Hinduphobic sentiment patterns, representing a significant advantage over single-task baseline models that require separate sentiment analysis pipelines.

\begin{table}[h]
\centering
\begin{tabular}{lcccc}
       \hline
Model & Accuracy & Precision & Recall & F1-Score \\
       \hline
DistilBERT \cite{sanh2019distilbert} & 0.8423 & 0.8398 & 0.8447 & 0.8423 \\      \hline
ELECTRA \cite{clark2020electra} & 0.8698 & 0.8675 & 0.8721 & 0.8698 \\      \hline
DeBERTa \cite{he2020deberta} & 0.8745 & 0.8723 & 0.8767 & 0.8745 \\      \hline
RoBERTa \cite{liu2019roberta} & 0.8812 & 0.8789 & 0.8835 & 0.8812 \\      \hline
HateBERT \cite{caselli2020hatebert} & 0.8901 & 0.8878 & 0.8924 & 0.8901 \\      \hline
HP-BERT (Ours) & 0.9472 & 0.9474 & 0.9467 & 0.9465 \\
       \hline
\end{tabular}
\caption{Fine-tuned model comparison on Hinduphobic tweets detection}
\label{tab:model_comparison_finetuned}
\end{table}



\subsection{Data Analysis on Global COVID-19 X (Twitter) Dataset)}

\subsubsection{Analysis of COVID-19 Cases}

\begin{figure*}[htbp!]
    \centering
    \includegraphics[width=0.9\linewidth]{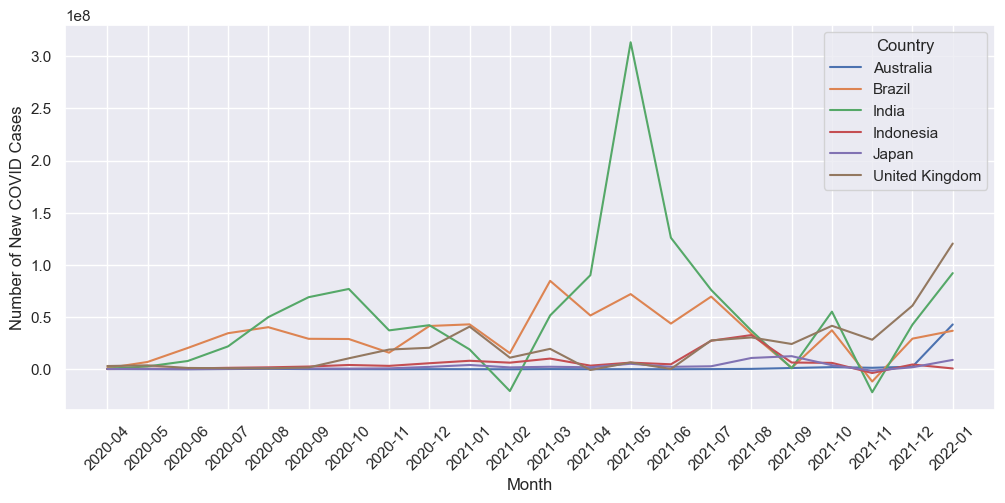}
    \caption{Monthly trends in COVID-19 case counts.}
    \label{Img.COVID_cases_over_months}
\end{figure*}

We first review the COVID-19 case trends from April 2020 to January 2022 across six countries, as shown in (Figure \ref{Img.COVID_cases_over_months}). We notice that India had a major peak in cases in April- May 2021, likely due to the Delta variant, while Brazil and Indonesia also saw significant peaks in mid-2021. The United Kingdom shows a rise toward late 2021, likely from the Omicron variant. Japan and Australia had relatively low case numbers throughout, with slight increases at the end of 2021. Overall, the trends reflect the varied impact of COVID-19 waves and variants spread across these regions.

\begin{figure*}[htbp!]
\small
    \centering
    \includegraphics[width=0.9\linewidth]{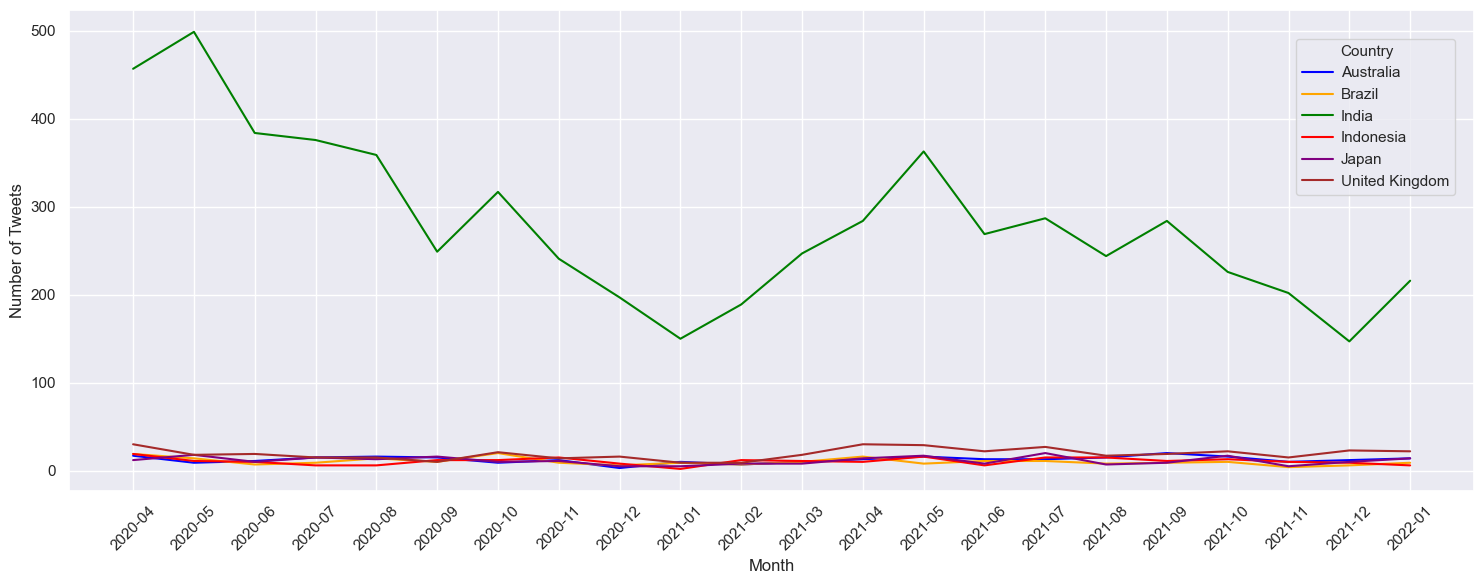}
    \caption{Monthly trends of Hinduphobic tweets across all countries.}
    \label{Img.All_Country_Number_of_Tweets}
\end{figure*}

\begin{figure*}[htbp!]
    \centering
    \includegraphics[width=0.9\linewidth]{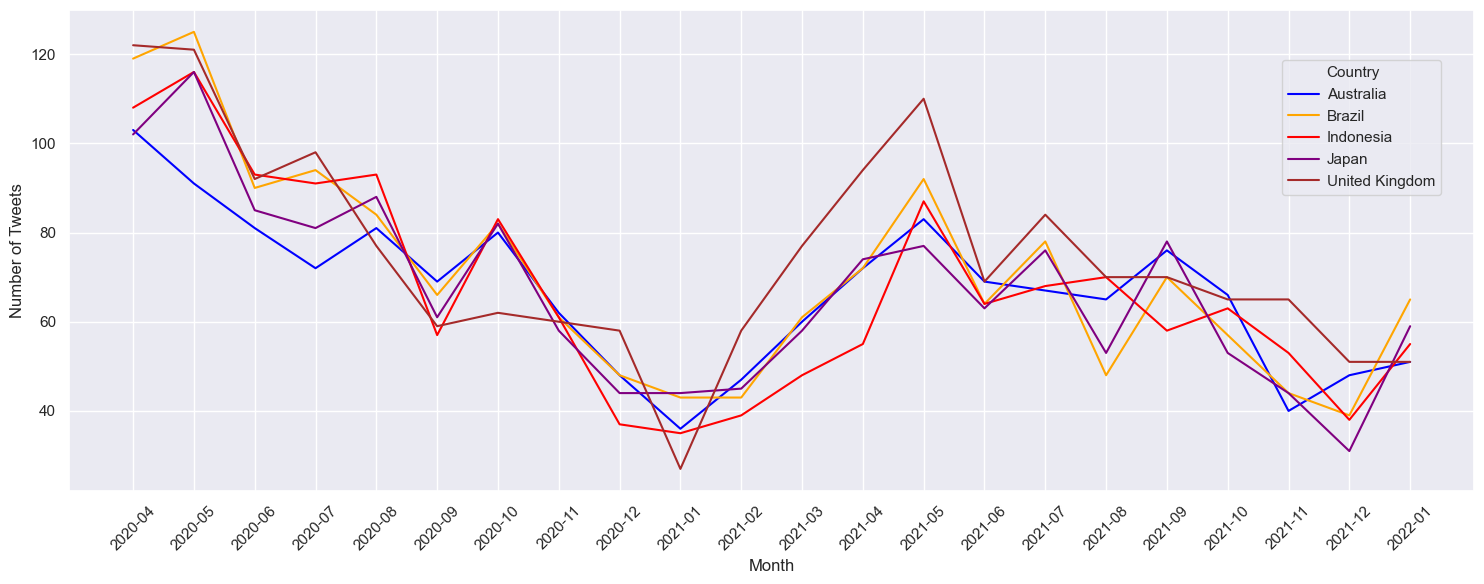}
    \caption{Monthly trends of Hinduphobic tweets across countries excluding India.}
    \label{Img.Excluding_India_Number_of_Tweets}
\end{figure*}


\begin{table*}[htbp!]
\centering
\resizebox{\textwidth}{!}{ 
\begin{tabular}{l|l|c|c|c|c|c|c}
\hline
\textbf{Dataset} & \textbf{Method} & \textbf{Australia} & \textbf{Brazil} & \textbf{India} & \textbf{Indonesia} & \textbf{Japan} & \textbf{UK} \\ \hline
\textbf{The Global Dataset} & HateBERT \cite{caselli2020hatebert} &  137699 & 36734 & 151659 & 8109 & 26518 & 64121  \\ \hline
\textbf{The Global Dataset} & HP-BERT & 479 & 352 & 9242 & 356 & 389 & 746 \\ \hline
\textbf{Subset of The Global Dataset} & HateBERT \cite{caselli2020hatebert} & 48 & 33 & 737 & 39 & 29 & 65 \\ \hline
\end{tabular}}
\caption{Tweet counts across six countries using HateBERT and HP-BERT on the global dataset and its subset.}
\label{tab:BERTresults}
\end{table*}

\subsubsection{Results of HateBERT model}

\begin{figure}[htbp!]
    \centering
    \begin{subfigure}[b]{0.45\textwidth}
        \includegraphics[width=\textwidth]{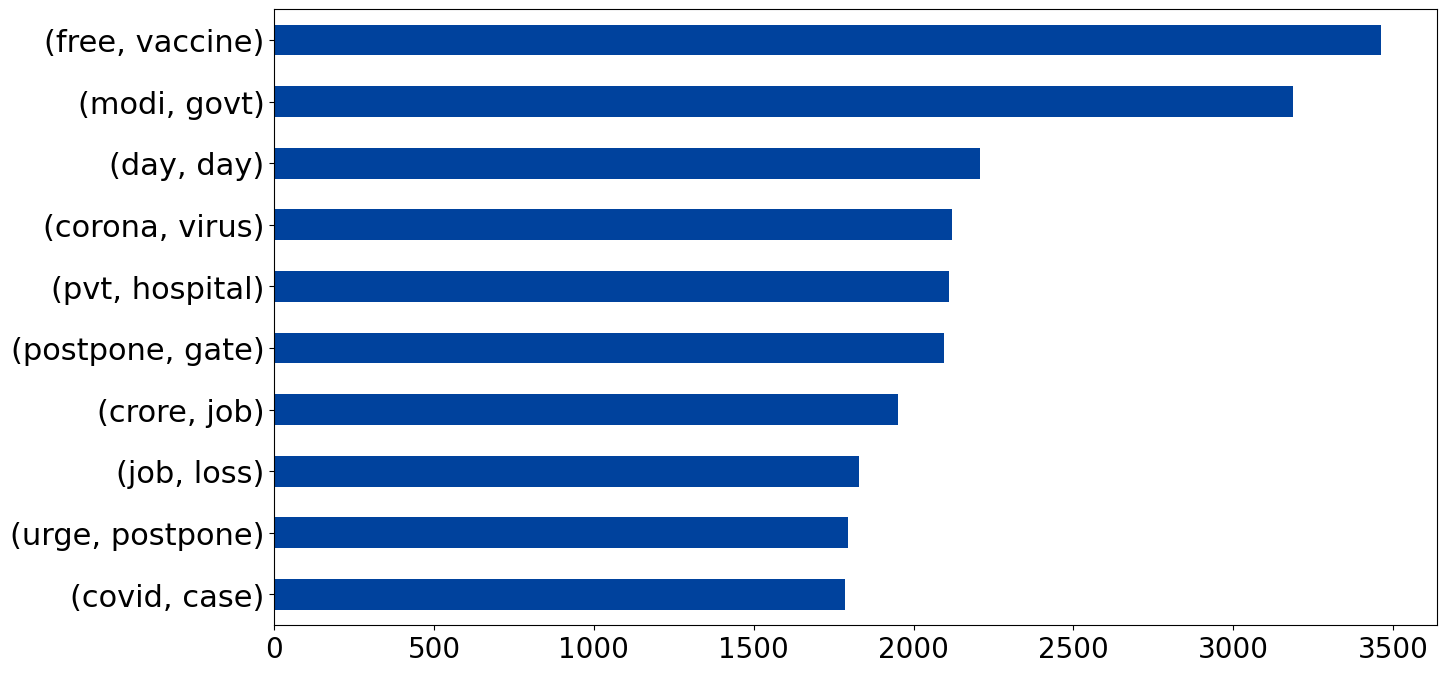}
        \caption{Bigram}
        \label{BH1}
    \end{subfigure}
    \hfill
    \begin{subfigure}[b]{0.45\textwidth}
        \includegraphics[width=\textwidth]{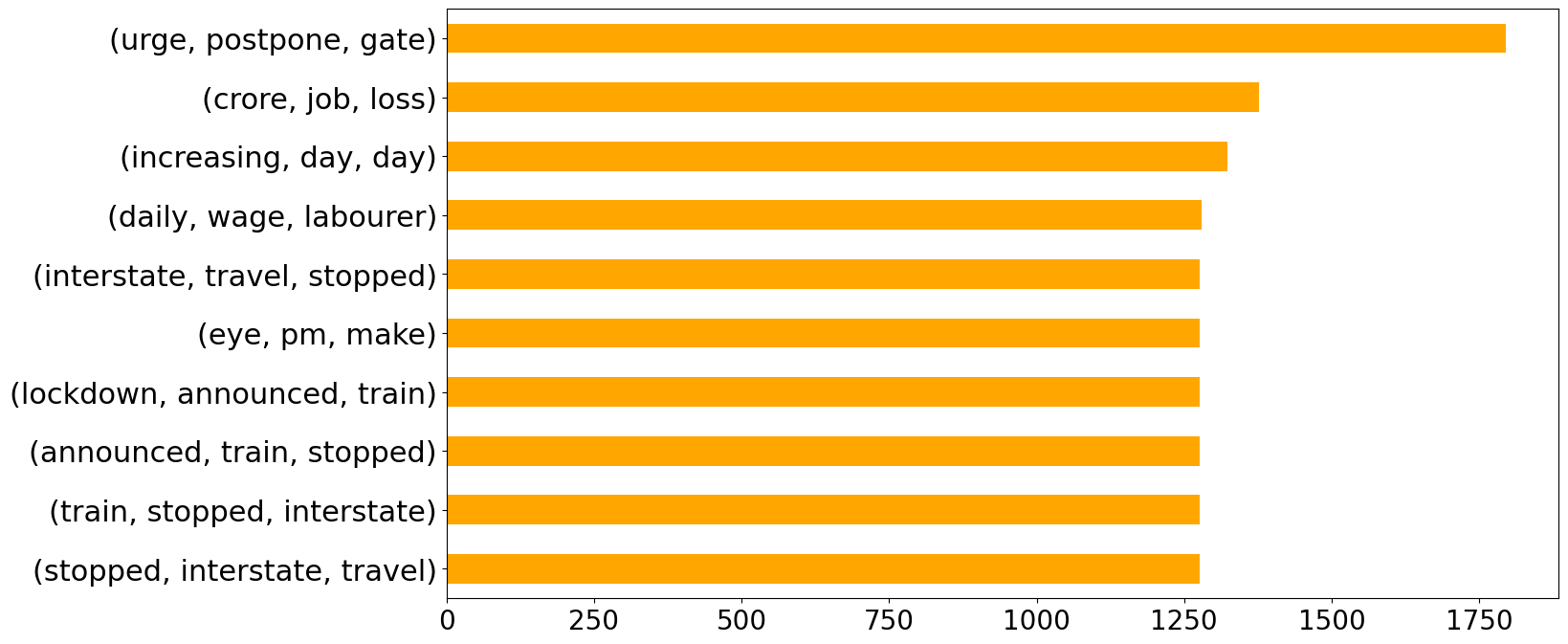}
        \caption{Trigram}
        \label{TH1}
    \end{subfigure}
    \caption{Bigram and trigram analysis of tweets labelled by HateBERT worldwide.}
    \label{HateBERTfig:bigram_trigram}
\end{figure}

We applied HateBERT \cite{caselli2020hatebert} to analyze hate speech and abusive content during the COVID-19 pandemic across six countries, with the results summarised in (Table \ref{tab:BERTresults}). To gain deeper insights into recurring themes within the discourse, we performed bigram and trigram analyses, as shown in (Figure \ref{HateBERTfig:bigram_trigram} Panel-(a)), which shows the most frequent bigrams, which reveal key patterns in the pandemic-related discourse. Bigrams such as "free vaccine" and "pvt hospital" highlight concerns around vaccine accessibility and healthcare disparities, while "modi govt" and "crore job" reflect criticisms of the Indian government and the economic challenges faced by citizens. (Figure \ref{HateBERTfig:bigram_trigram} Panel-(b)) presents the most frequent trigrams, offering further insights into more complex interactions within the conversations. Trigrams such as "daily wage labourer" and "crore job loss" underscore the economic vulnerabilities faced by workers during the pandemic, while "lockdown announced train" and "stopped interstate travel" emphasise the disruptions caused by lockdown measures, particularly on mobility and commerce.  

Although the results from HateBERT provide valuable general insights into pandemic-related discourse, they remain broad. To further refine our analysis and specifically focus on Hinduphobic content, we applied the HP-BERT model in subsequent analyses.  

\subsubsection{Results HP-BERT model}

Using the HP-BERT model, we analyze Hindubhobic tweets and associated sentiments across countries and regions, providing insight into the perceptions of Hindu communities in various parts of the world during and after the COVID-19 pandemic. This can enable us to identify sentiment trends and potentially inform strategies for addressing harmful stereotypes and communal biases on social media platforms.

\begin{figure}[htbp!]
    \centering
    \includegraphics[width=1\linewidth]{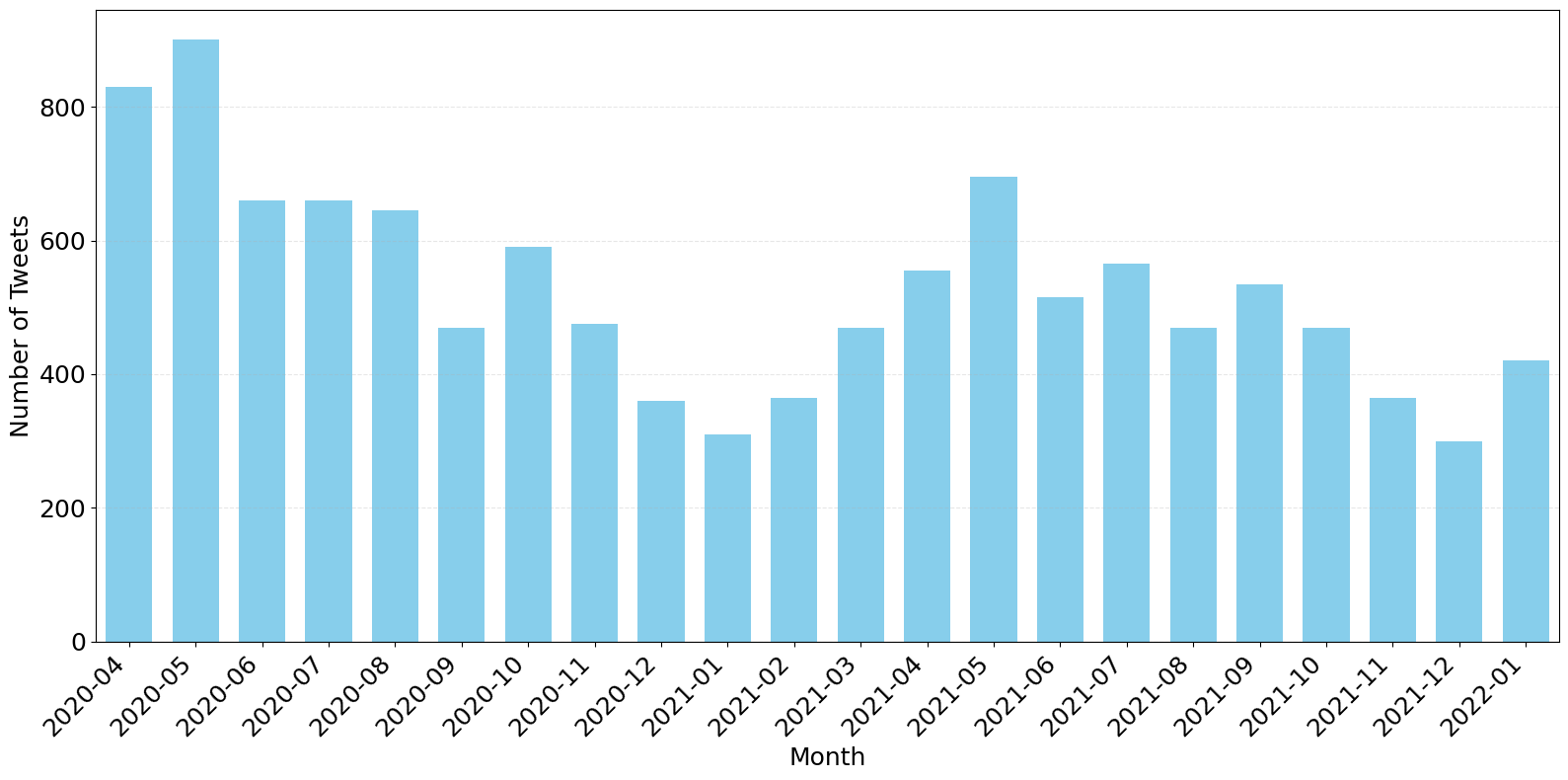}
\caption{Monthly distribution of Hinduphobic tweets identified by the HP-BERT model.}
    \label{Img.Number_of_tweets_monthwise}
\end{figure}

Finally, we applied our HP-BERT model to the Global COVID-19 X (Twitter) Dataset to analyze Hinduphobic tweets. (Figure \ref{Img.All_Country_Number_of_Tweets}) presents a multi-graph chart illustrating tweet counts across different countries, showing the monthly distribution of Hinduphobic tweet counts from April 2020 to January 2022. Table \ref{tab:BERTresults} provides a detailed breakdown of the tweet counts across six countries, where India recorded the highest number of Hinduphobic tweets, accounting for over 9,242 tweets, significantly surpassing other countries such as the United Kingdom, Australia, Japan, Indonesia, and Brazil, which reported much lower tweet counts. This higher number in India can be attributed to its large Hindu population \cite{scaccianoce2021people}, along with the presence of other significant religious communities, such as Muslims and Christians. The United Kingdom ranked second, followed by Australia, Japan, Indonesia, and Brazil. This analysis highlights that the Hinduphobic sentiment on Twitter during this period was predominantly concentrated in India. According to the 2011 Indian census \cite{chandramouli2011census}, Hinduism encompassed 79.8\% of the population, followed by 14.2\% Islam and 2.3\% Christianity.  
 
We created an additional visualization excluding India, as shown in (Figure \ref{Img.Excluding_India_Number_of_Tweets}) to allow a clearer view of Hinduphobic tweet counts in the remaining countries. We also present a distribution of Hinduphobic tweets (Figure \ref{Img.Number_of_tweets_monthwise}) to review the fluctuating levels of Hinduphobic tweets, with notable peaks (April to May 2020 and April to May 2021). After these peaks, we can observe a general downward trend with periodic increases from March to April 2021 and September 2021. This distribution suggests varying intensities of Hinduphobic sentiment on Twitter throughout the observed period, with some months experiencing sharp rises in tweet volume while others show a noticeable decline.

\begin{figure}[htbp!]
    \centering
    \begin{subfigure}[b]{0.49\textwidth}
        \includegraphics[width=\textwidth]{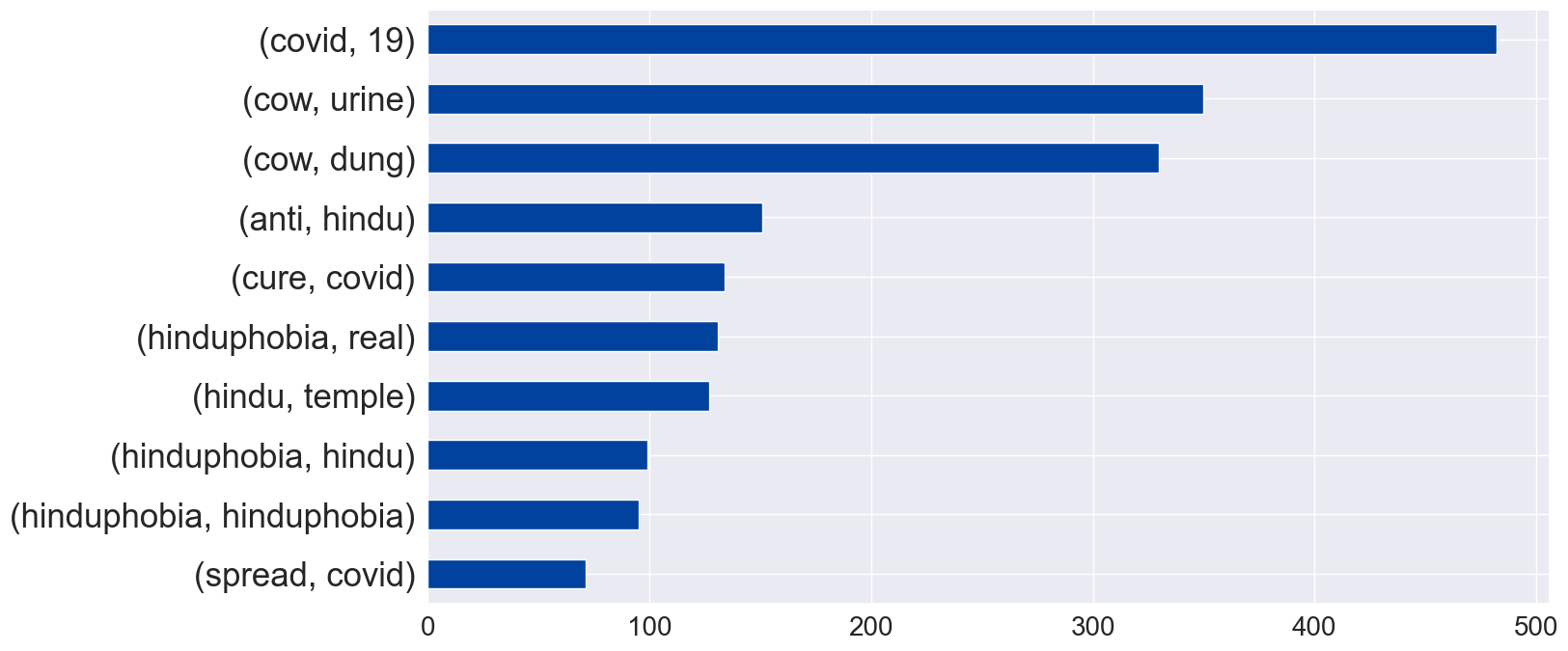}
        \caption{Bigram}
        \label{BH2}
    \end{subfigure}
    \hfill
    \begin{subfigure}[b]{0.49\textwidth}
        \includegraphics[width=\textwidth]{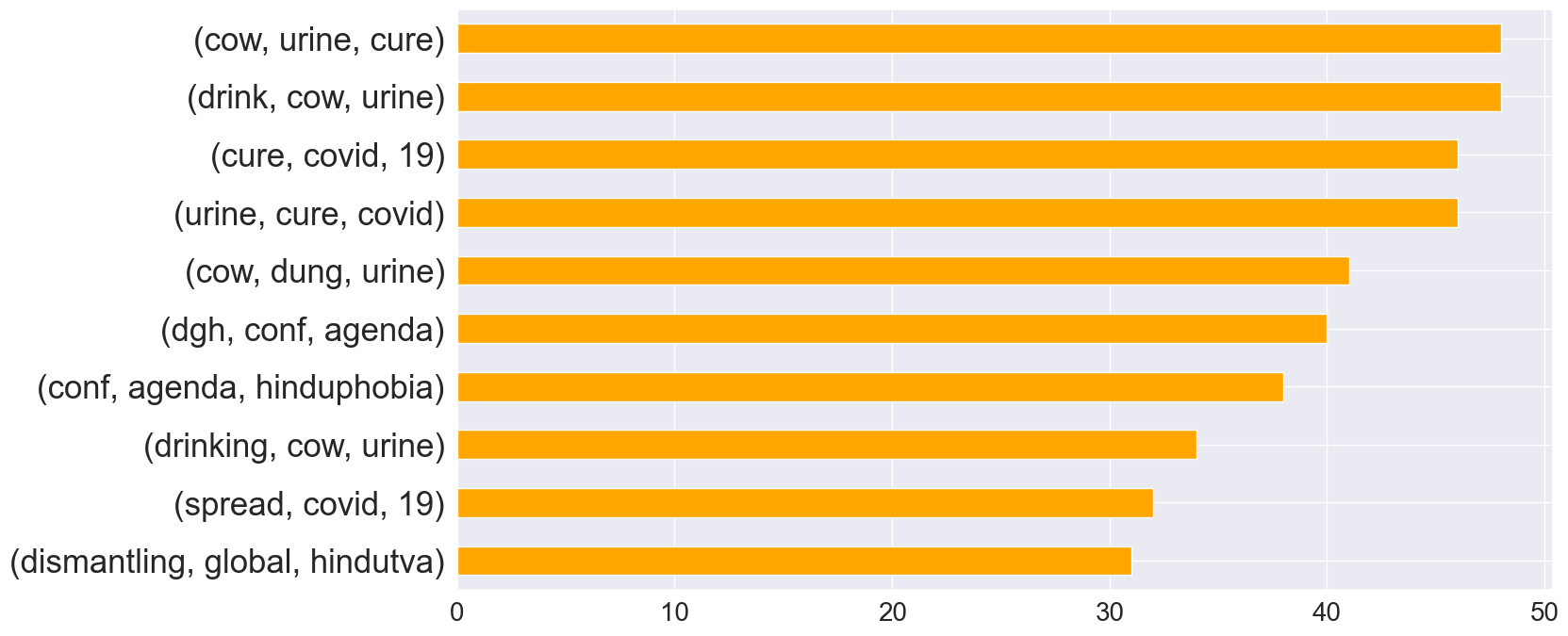}
        \caption{Trigram}
        \label{TH2}
    \end{subfigure}
\caption{Bigrams and trigrams extracted from Hinduphobic tweets identified by HP-BERT worldwide.}
    \label{fig:HP-BERTbigram_trigram}
\end{figure}

We performed n-gram analysis \cite{brown1992class} to explore the tweets processed by HP-BERT, focusing on bigrams and trigrams. (Figure \ref{fig:HP-BERTbigram_trigram}) illustrates the top 10 bigrams and trigrams extracted from tweets posted between April 2020 and January 2022 across six countries. In (Figure \ref{fig:HP-BERTbigram_trigram} Panel-(a)), the most frequent bigrams in the dataset include "covid 19", "cow urine" and "cow dung". Notably, the bigram "cure covid" appears frequently, suggesting widespread discussions mocking Hindus for purportedly using cow urine and cow dung as COVID-19 treatment due to the sacred nature of cows in Hinduism. Other bigrams, such as "anti hindu", "hindu temple" and "spread covid" highlight tweets blaming Hindus for spreading COVID-19 by attending temples. Furthermore, phrases such as "hinduphobia hindu", "hinduphobia hinduphobia" and "hinduphobia real" contribute to the growing narrative of negative sentiment against Hindus.

(Figure \ref{fig:HP-BERTbigram_trigram} Panel-(b)) presents the most frequent trigrams, including "cow urine cure", "drink cow urine", "urine cure covid", "cow dung urine" and "cure covid 19". These terms reflect stereotypes about Hindus using cow-related products to combat COVID-19, as seen in reports of events such as cow urine-drinking gatherings to ward off the virus \cite{daria2021use} and such narratives have contributed to discrimination against Hindus in several countries \cite{saiya2024faith}. Additionally, trigrams like "dgh conf agenda" and "conf agenda hinduphobia" refer to a conference held in the United States aimed at addressing global Hinduphobia\footnote{\url{https://shorturl.at/8xCUW}}. Phrases such as "spread covid 19" and "dismantling global hindutva" explicitly link Hindus to the spread of COVID-19, further fuelling anti-Hindu sentiments worldwide.

\begin{figure*}[htbp] 
\begin{subfigure}{0.48\textwidth}
\includegraphics[width=\linewidth]{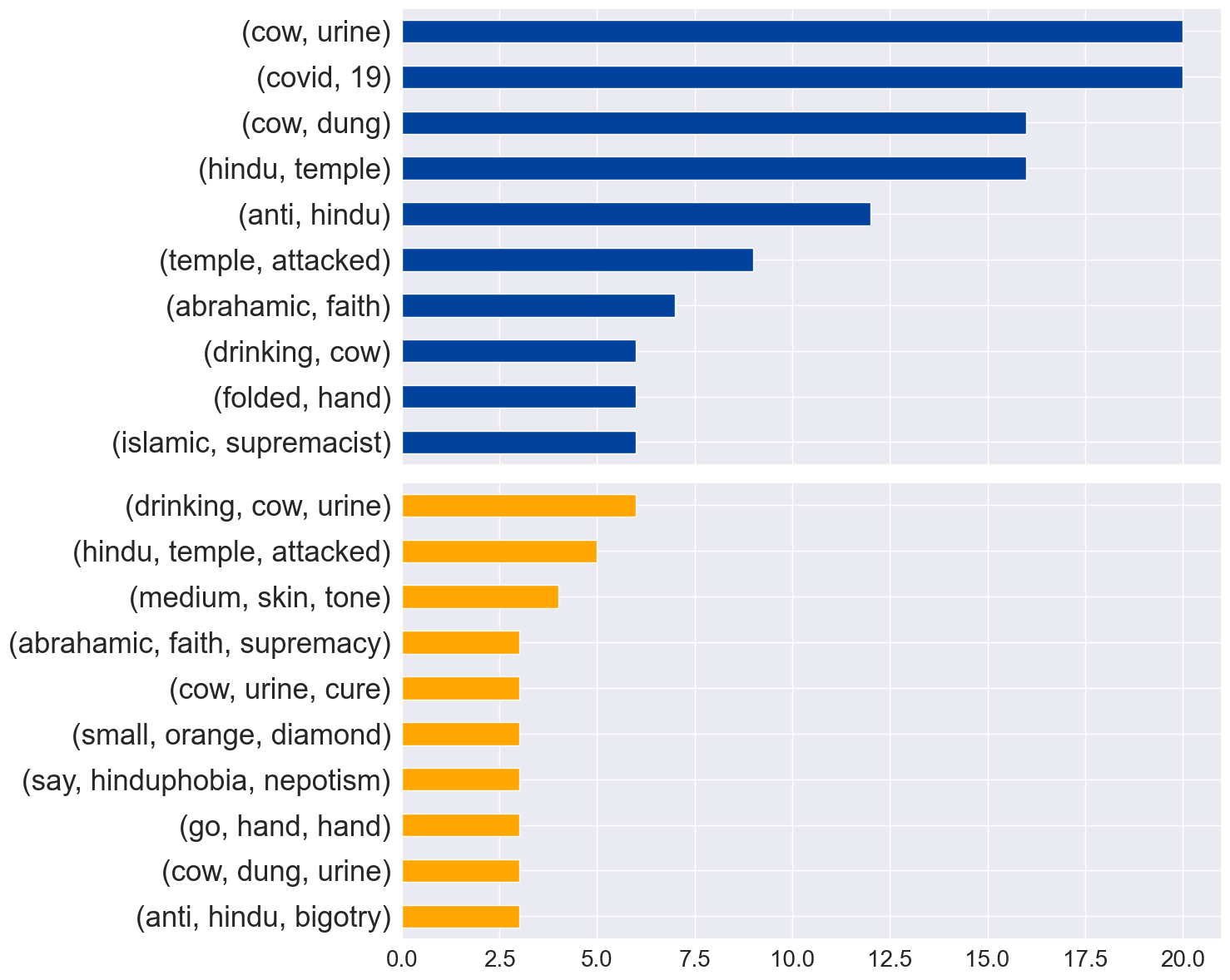}
\caption{Australia} \label{fig:a}
\end{subfigure}\hspace*{\fill}
\begin{subfigure}{0.48\textwidth}
\includegraphics[width=\linewidth]{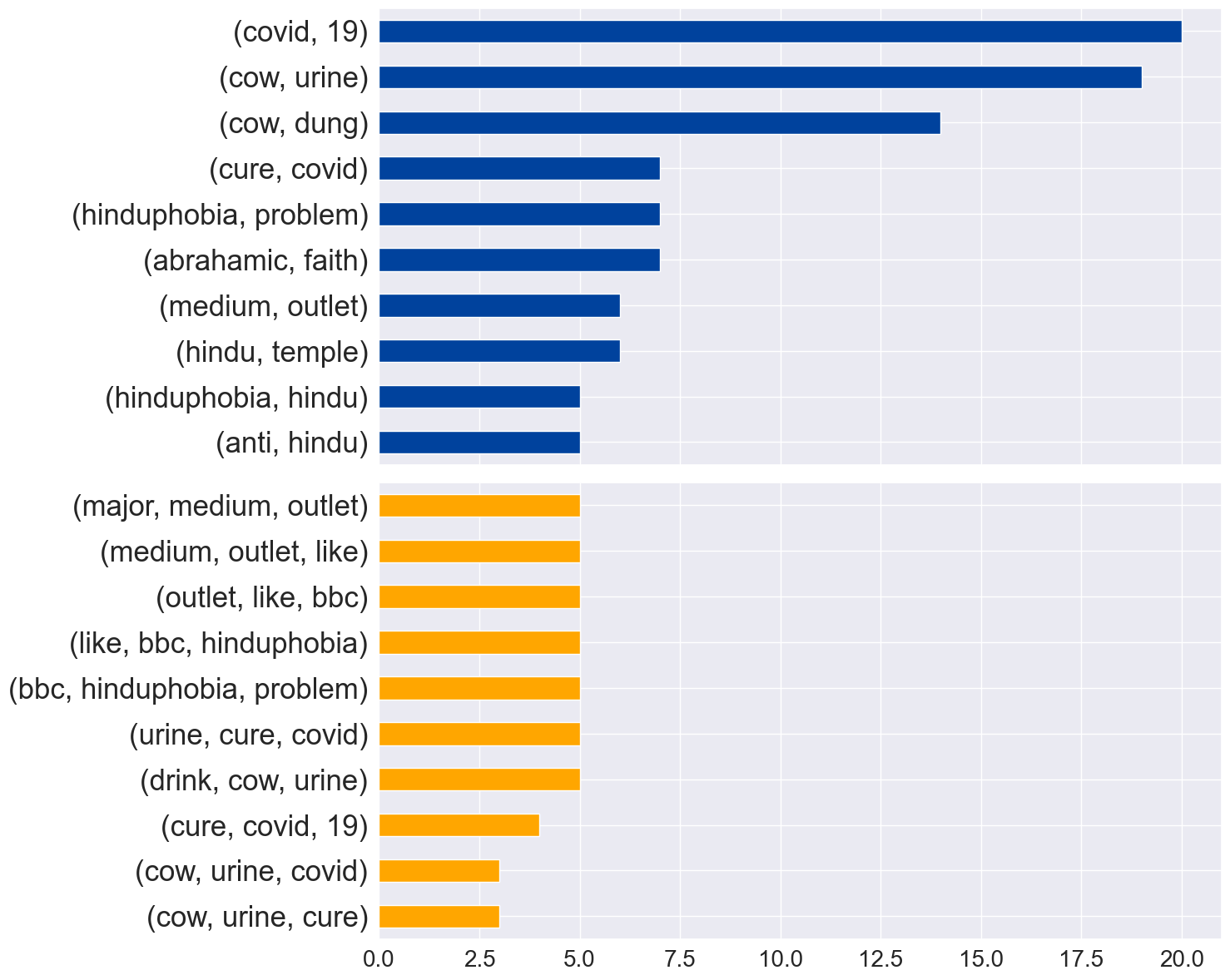}
\caption{Brazil} \label{fig:b}
\end{subfigure}
\medskip
\begin{subfigure}{0.48\textwidth}
\includegraphics[width=\linewidth]{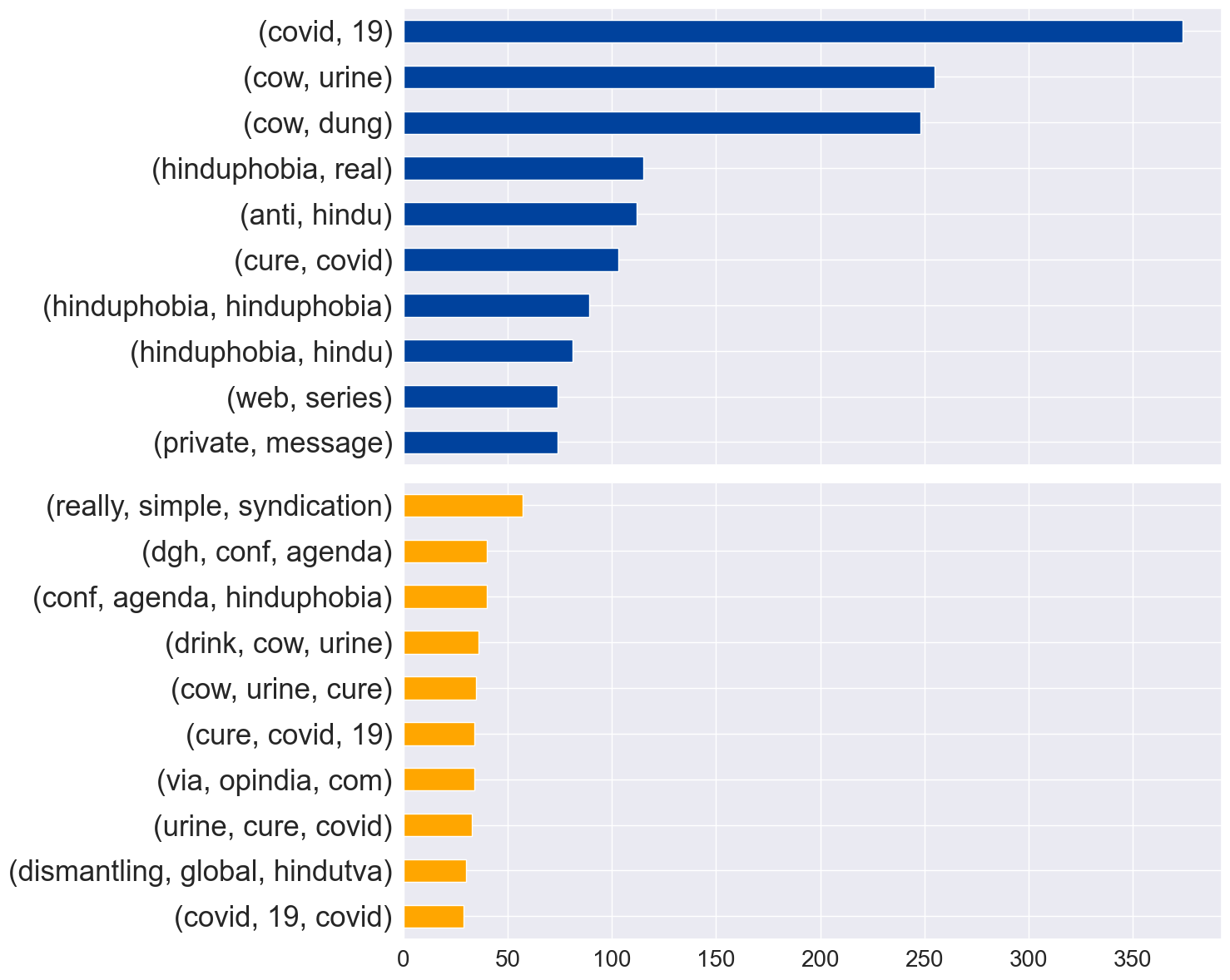}
\caption{India} \label{fig:c}
\end{subfigure}\hspace*{\fill}
\begin{subfigure}{0.48\textwidth}
\includegraphics[width=\linewidth]{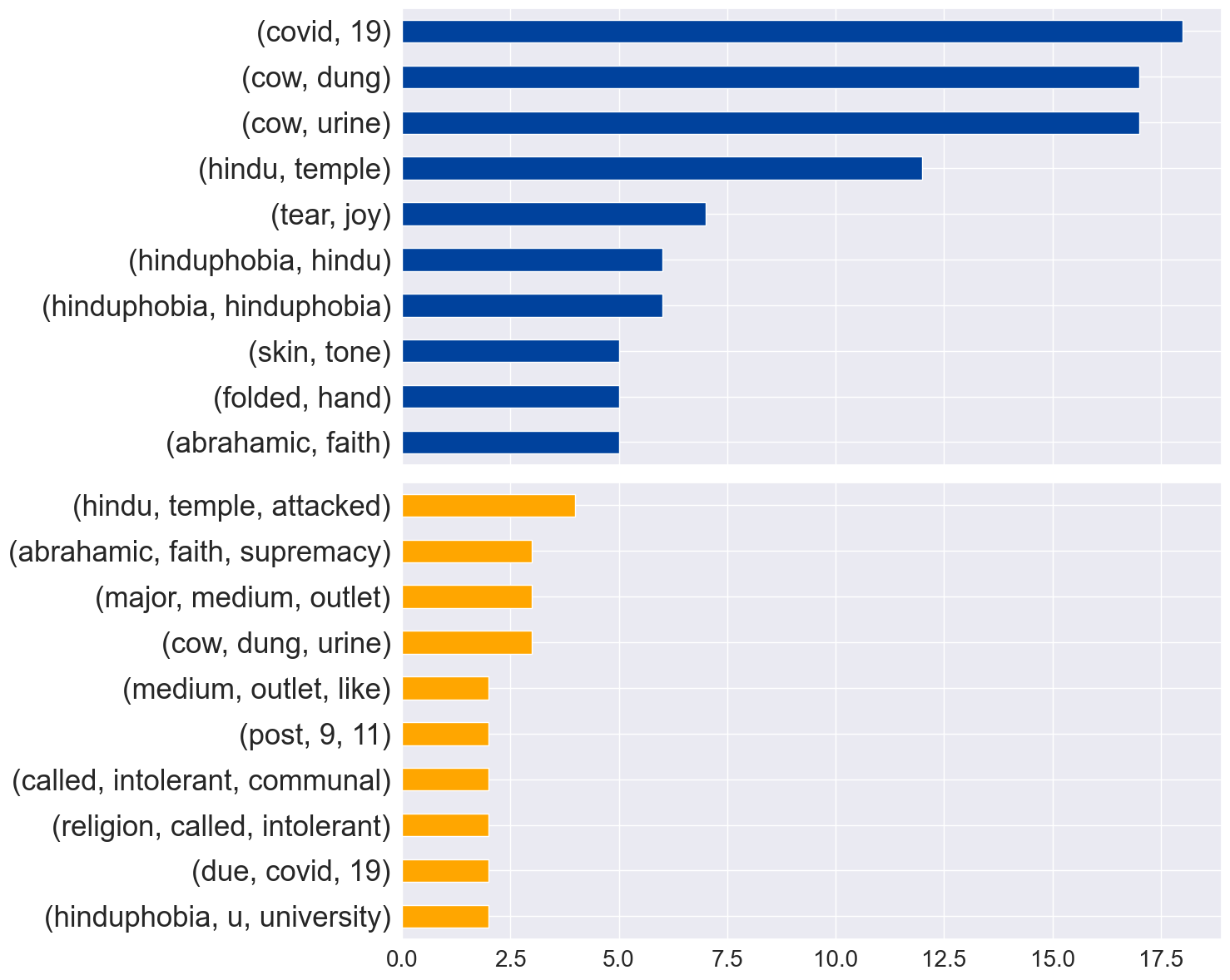}
\caption{Indonesia} \label{fig:d}
\end{subfigure}
\medskip
\begin{subfigure}{0.48\textwidth}
\includegraphics[width=\linewidth]{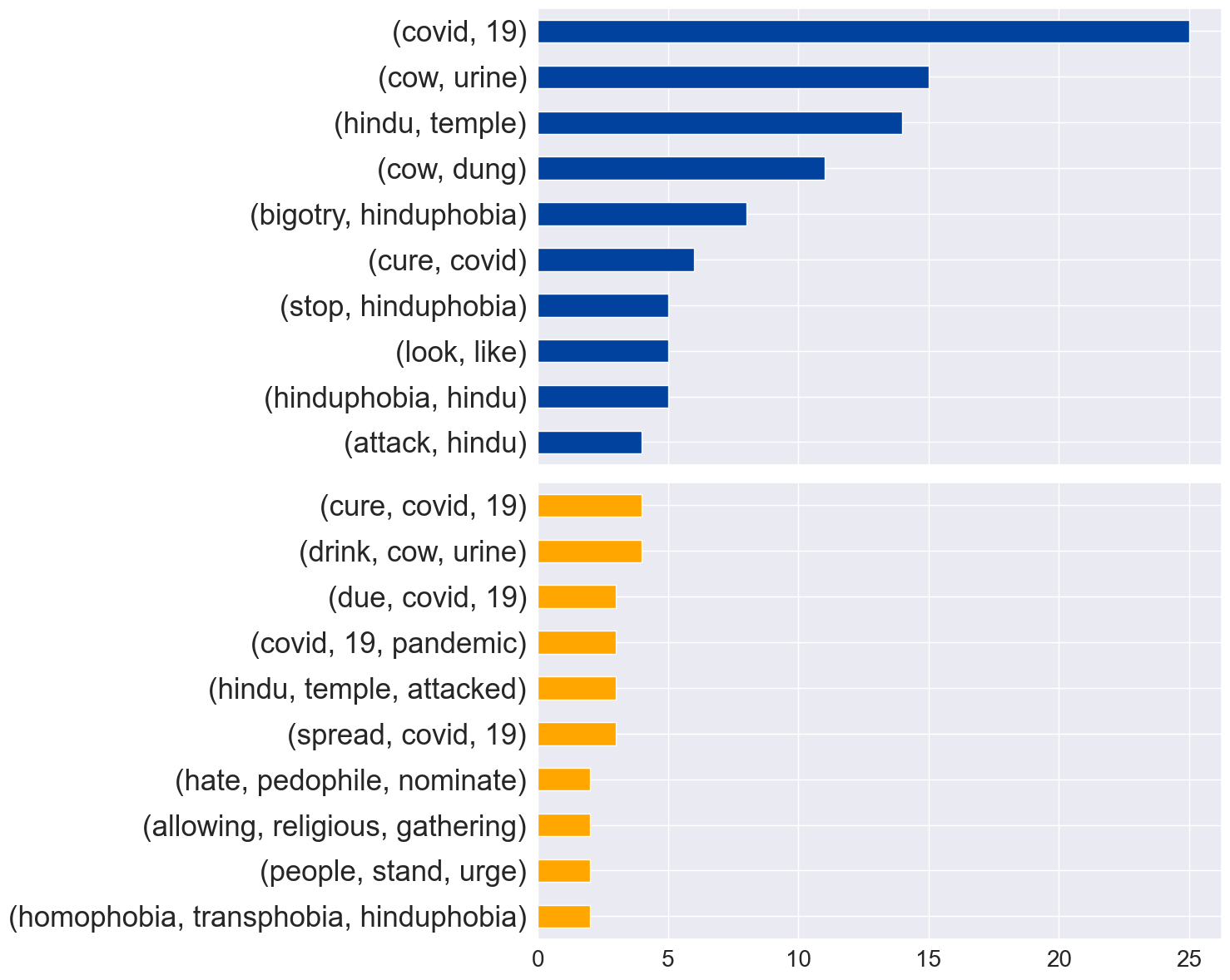}
\caption{Japan} \label{fig:e}
\end{subfigure}\hspace*{\fill}
\begin{subfigure}{0.48\textwidth}
\includegraphics[width=\linewidth]{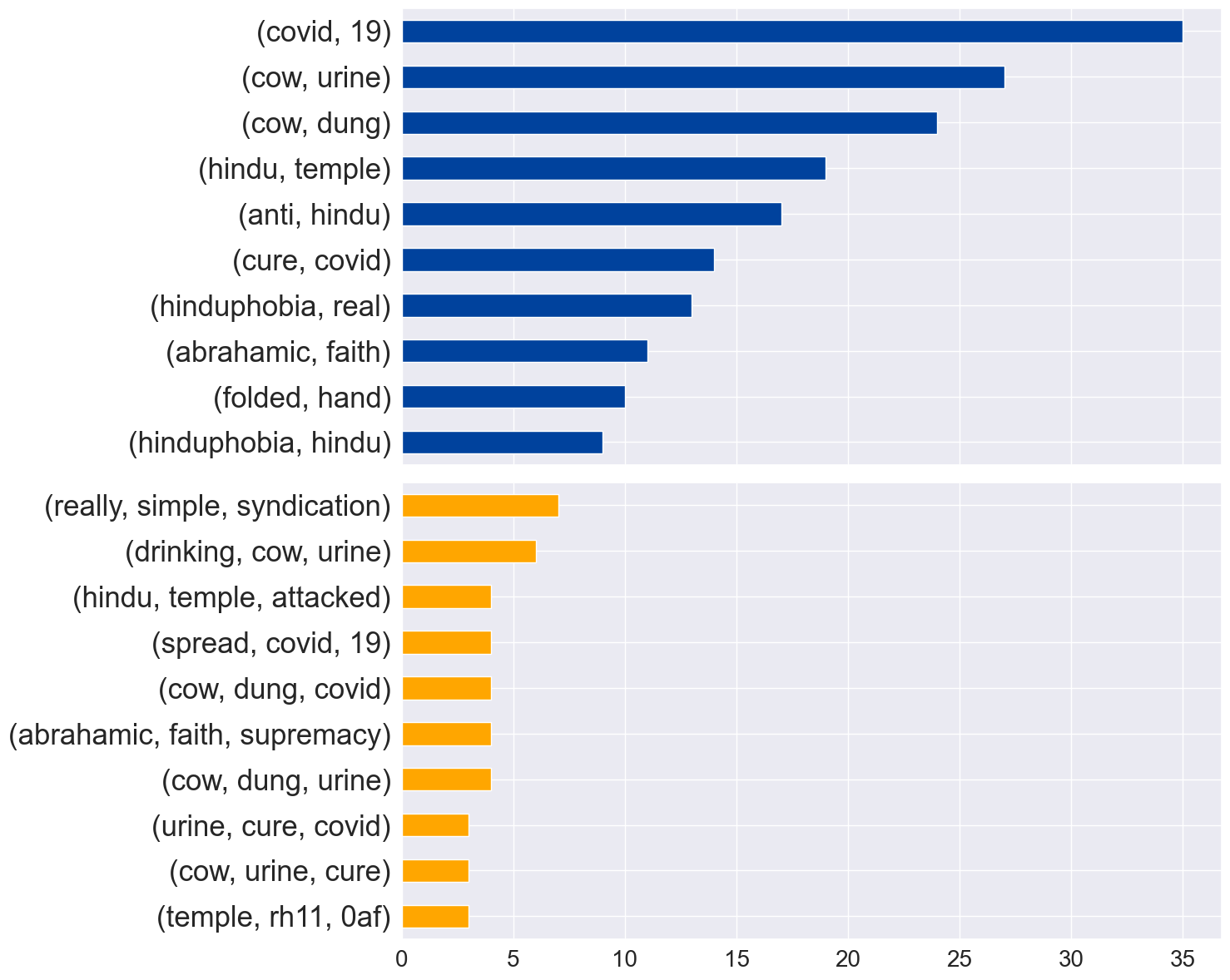}
\caption{United Kingdom} \label{fig:f}
\end{subfigure}
\caption{Top 10 bigrams and trigrams of Hinduphobic tweets identified by HP-BERT for different countries.}
\label{fig:HPBert_bigraph&trigram_thumbnails}
\end{figure*}

We also extracted the top 10 bigrams and trigrams for each of the six countries, as shown in (Figure \ref{fig:HPBert_bigraph&trigram_thumbnails}). We find that certain bigrams, such as "covid 19", "cow urine" and "cow dung" are consistently present across all six countries. Similarly, the most frequent trigrams are common across countries, aligning with global patterns. These recurring phrases reflect a shared narrative that transcends national boundaries, emphasising the widespread dissemination of stereotypes and sentiments linked to Hinduphobia.

In Australia, notable bigrams include "hindu temple", "hinduphobia hindu", "anti hindu", "folded hand" (a reference to the Indian greeting "Namaste"), and "abrahamic faith" often used to mock Hindu practices and promote a sense of religious superiority. Trigrams such as "hindu temple attacked", "anti hindu bigotry" and "abrahamic faith supremacy" highlight themes of discrimination and religious intolerance. In Brazil, key bigrams include "cure covid", "hinduphobia problem", "abrahamic faith", and "hindu temple". The trigrams "like bbc hinduphobia", "bbc hinduphobia problem", "urine cure covid", "drink cow urine" and "cure covid 19" indicate mockery directed toward Hindu practices, particularly regarding traditional remedies. In India, prominent bigrams include "hinduphobia real", "anti hindu", and "hinduphobia hindu". Trigrams such as "drink cow urine", "cow urine cure", "cure covid 19", and "dismantling global hindutva" underscore narratives aimed at ridiculing Hindu practices and beliefs. For Indonesia, bigrams such as  "hindu temple", "hinduphobia hindu", "skin tone" and "folded hand" point to cultural and racial stereotyping. The trigrams such as "hindu temple attacked", "abrahamic faith supremacy", "cow dung urine" and "religion called intolerant" suggest mockery of Hindu traditions and communal narratives. In Japan, bigrams such as  "cure covid", "stop hinduphobia", and "attack hindu" reflect growing concerns about Hinduphobia. The trigrams such as "cure covid 19", "drink cow urine", "hindu temple attacked" and "spread covid 19" emphasise the global spread of derogatory narratives targeting Hindus. Lastly, in the United Kingdom, bigrams like "hindu temple", "anti hindu" and "hinduphobia real" are significant. Trigrams such as "drinking cow urine", "hindu temple attacked", "spread covid 19" and "cow dung urine" further illustrate how stereotypes are perpetuated. Overall, while common themes of Hinduphobia are observed across countries, specific regional narratives and stereotypes are evident in the analysis. These findings reveal how Hindu practices, beliefs, and culture are frequently mocked or misrepresented, reflecting shared global patterns and localised biases.

\subsection{Analysis of HP-BERT results using HateBERT}

\begin{figure}[htbp!]
    \centering
    \begin{subfigure}[b]{0.49\textwidth}
        \includegraphics[width=\textwidth]{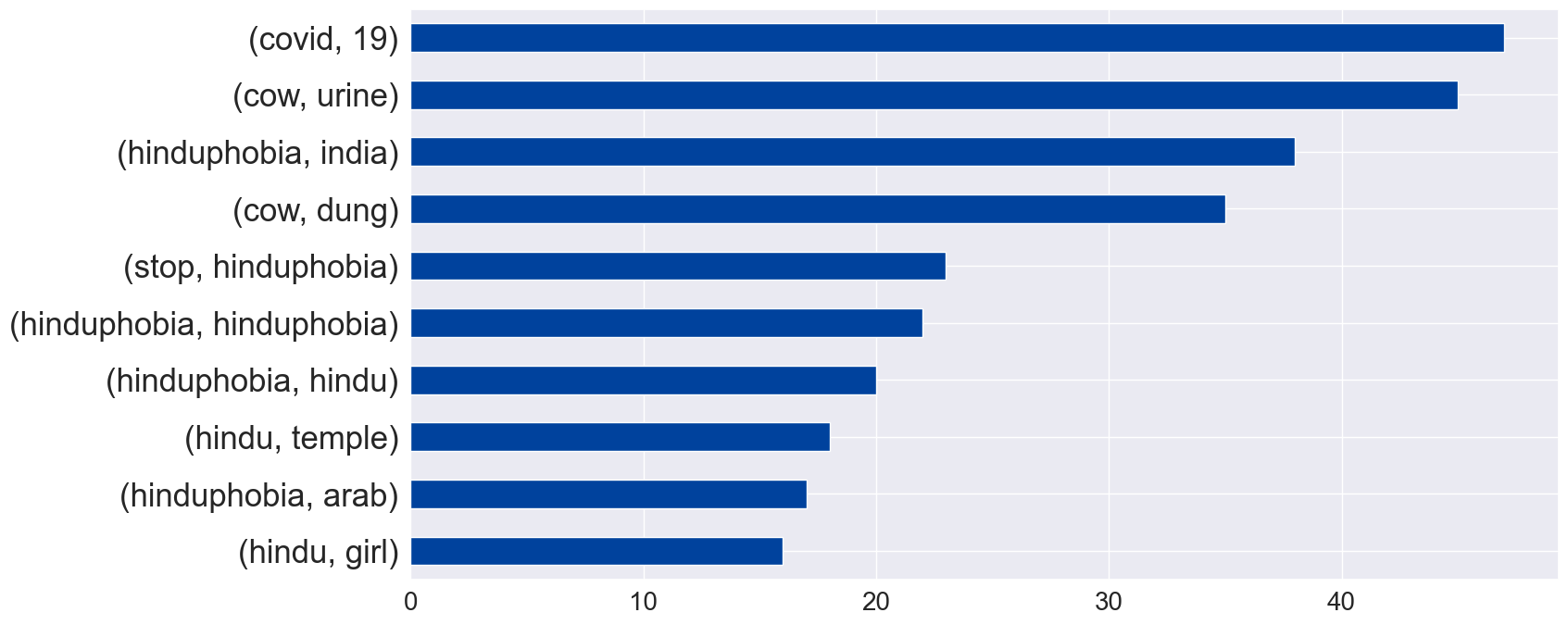}
        \caption{Bigram}
        \label{Bigram}
    \end{subfigure}
    \hfill
    \begin{subfigure}[b]{0.49\textwidth}
        \includegraphics[width=\textwidth]{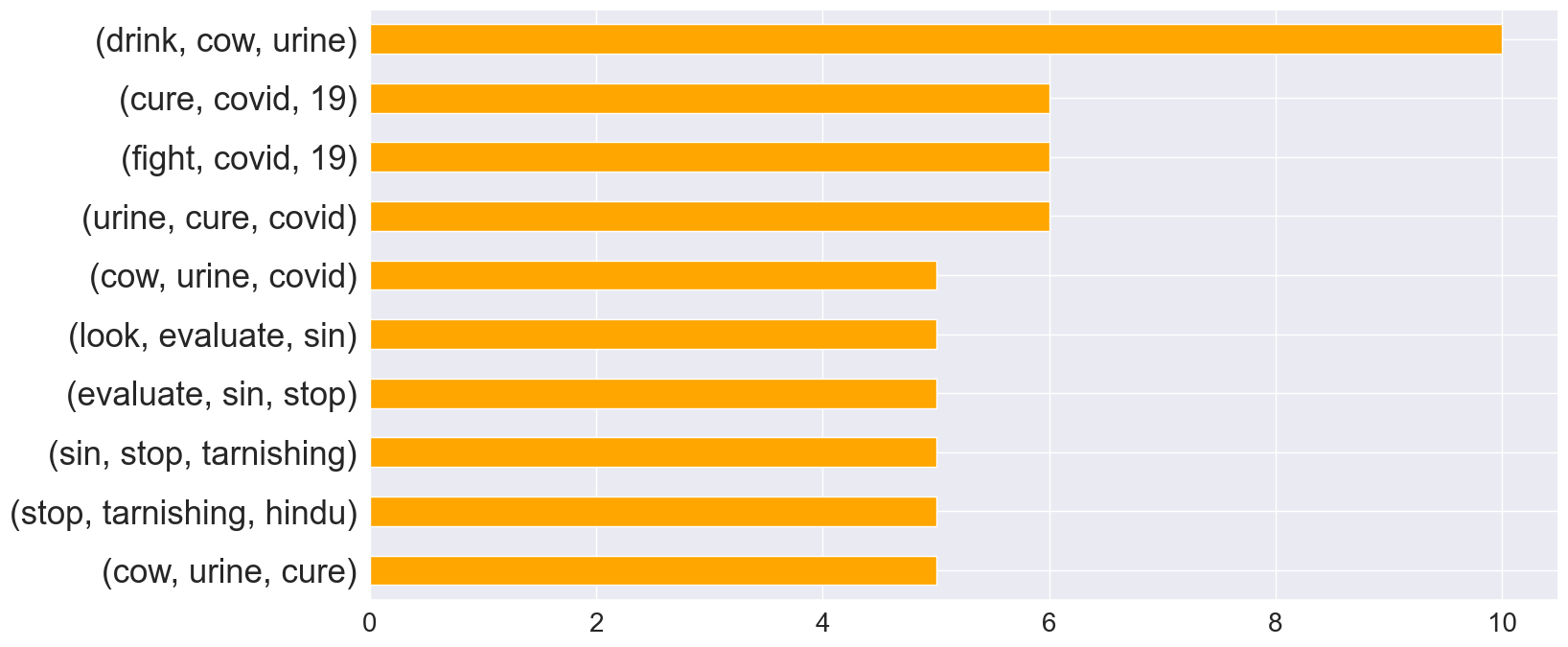}
        \caption{Trigram}
        \label{Trigram}
    \end{subfigure}
    \caption{Bigrams and trigrams of Hinduphobic tweets identified by HP-BERT analyzed using HateBERT worldwide.}
    \label{fig:HPHATEbigram_trigram}
\end{figure}

Next, we apply  HateBERT \cite{caselli2020hatebert} to the results of HP-BERT to assess its effectiveness in detecting hate speech and abusive content, particularly in the context of Hinduphobia across six countries (Table \ref{tab:BERTresults}). The detection results were comparatively lower because HateBERT is primarily trained on English language data, which may not effectively capture nuanced and culturally specific Hinduphobic rhetoric. HateBERT was pre-trained on a corpus of Reddit comments, making it more effective at detecting overtly hateful and abusive content. Furthermore, we generate bigrams and trigrams as shown in (Figure \ref{fig:HPHATEbigram_trigram}) to highlight recurring themes. These phrases reflect recurring themes that highlight narratives of Hinduphobia, particularly in the context of COVID-19 remedies, cow-related references, and religious or cultural stigmas. This analysis underscores the prevalence of stereotypes and how they manifest across different countries.

\subsection{Statistical Significance of Temporal and Geographical Patterns}

Our statistical analysis of the results confirms significant variations in Hinduphobic discourse across both temporal and geographical dimensions. The cross-country comparison using one-way Analysis of Variance (ANOVA) revealed significant differences in tweet volumes (F(5,156) = 18.67, p < 0.001), with India accounting for 84.7\% of total Hinduphobic content (9,242 tweets), significantly higher than the UK (746 tweets), Australia (479 tweets), Japan (389 tweets), Brazil (352 tweets), and Indonesia (356 tweets) as shown in (Table~\ref{tab:BERTresults}) (Tukey's Honestly Significant Difference (HSD), all p < 0.001). 

Temporal analysis using the Kruskal-Wallis test showed significant differences across pandemic phases (H = 12.45, p = 0.002). The during-COVID period (April 2020-December 2021) exhibited significantly higher tweet volumes (Mean = 142.3 tweets/month, Standard Deviation (SD) = 89.7) compared to pre-COVID baseline (January-March 2020: M = 67.2 tweets/month, SD = 34.5) and the post-peak period (January 2022 onwards: Mean = 98.4 tweets/month, SD = 52.1), as illustrated in (Figure~\ref{Img.COVID_cases_over_months}) and (Figure~\ref{Img.All_Country_Number_of_Tweets}). The results of our linear regression model revealed a significant increase trend during the study period for India ($\beta$ = 3.42, p = 0.018, R² = 0.23) and the UK ($\beta$ = 1.87, p = 0.041, R² = 0.16), confirmed by Mann-Kendall trend analysis (India: $\tau$ = 0.234, p = 0.012; UK: $\tau$ = 0.189, p = 0.038). 

Correlation analysis between monthly COVID-19 case counts (Figure~\ref{Img.Number_of_tweets_monthwise}) and Hinduphobic content revealed moderate positive associations across countries: India (r = 0.428, p = 0.016), UK (r = 0.391, p = 0.029), Australia (r = 0.312, p = 0.087), with weaker correlations in other countries. The peak coincidence analysis showed that 68\% of the major spikes in Hinduphobic content (defined as> 2 SD above mean) occurred within 4-6 weeks of the COVID-19 surge periods, significantly higher than expected by chance ($\chi^2$ = 6.73, p = 0.034). These findings, supported by the bigram and trigram analyses  (Figure~\ref{fig:HP-BERTbigram_trigram}) and (Figure~\ref{fig:HPBert_bigraph&trigram_thumbnails}), provide robust statistical evidence for pandemic-driven amplification of anti-Hindu sentiment, particularly in countries with larger Hindu diaspora populations.

\subsection{HP-BERT Sentiment Analysis}
We next employ the HP-BERT to perform sentiment analysis on Hinduphobic tweets, capturing a range of sentiment categories. (Table \ref{tab:key_evaluation_metrics}) provides evaluation metrics such as accuracy, precision, recall, and F1-score. Our analysis reveals a wide range of sentiments expressed in the tweets. We also examined user networks, activity patterns, and profile descriptions to identify common traits among users sharing Hinduphobic content. These insights offer a deeper understanding of the context and spread of Hinduphobia across social media platforms.

\begin{figure}[htbp!]
\small
    \centering
    \includegraphics[width=0.4\textwidth]{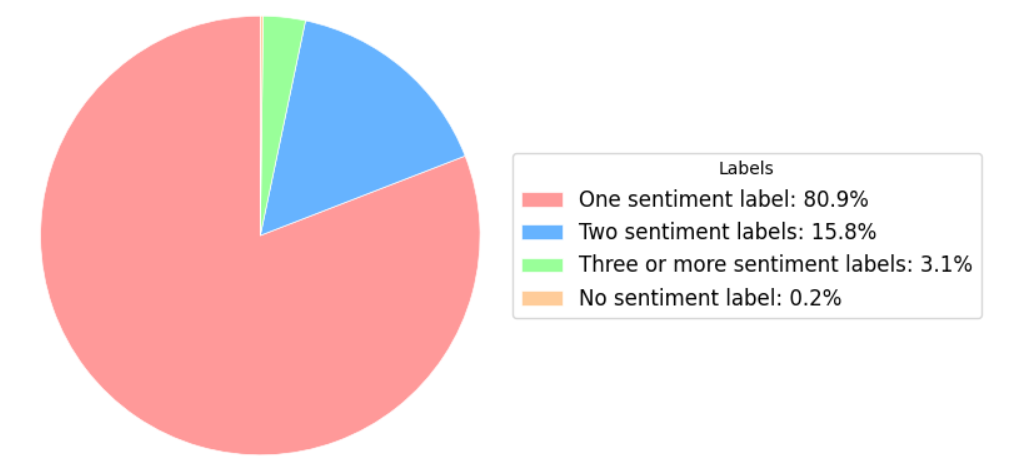}
     \caption{Percentage of Hinduphobic tweets with a different number of labels.}
    \label{fig:pielabels}
\end{figure}

\begin{figure}[htbp!]
\small
    \centering
    \includegraphics[width=0.45\textwidth]{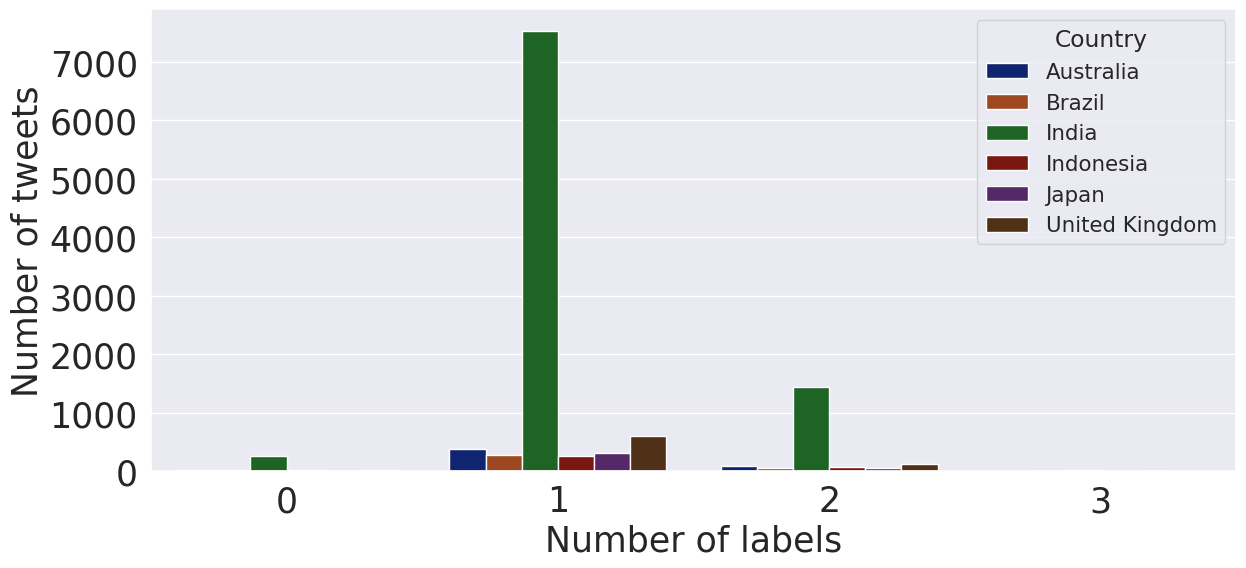}
    \caption{Number of Hinduphobic tweets with different numbers of labels}
    \label{fig:barlables}
\end{figure}

\begin{figure}[htbp!]
    \centering
    \includegraphics[width=1\linewidth]{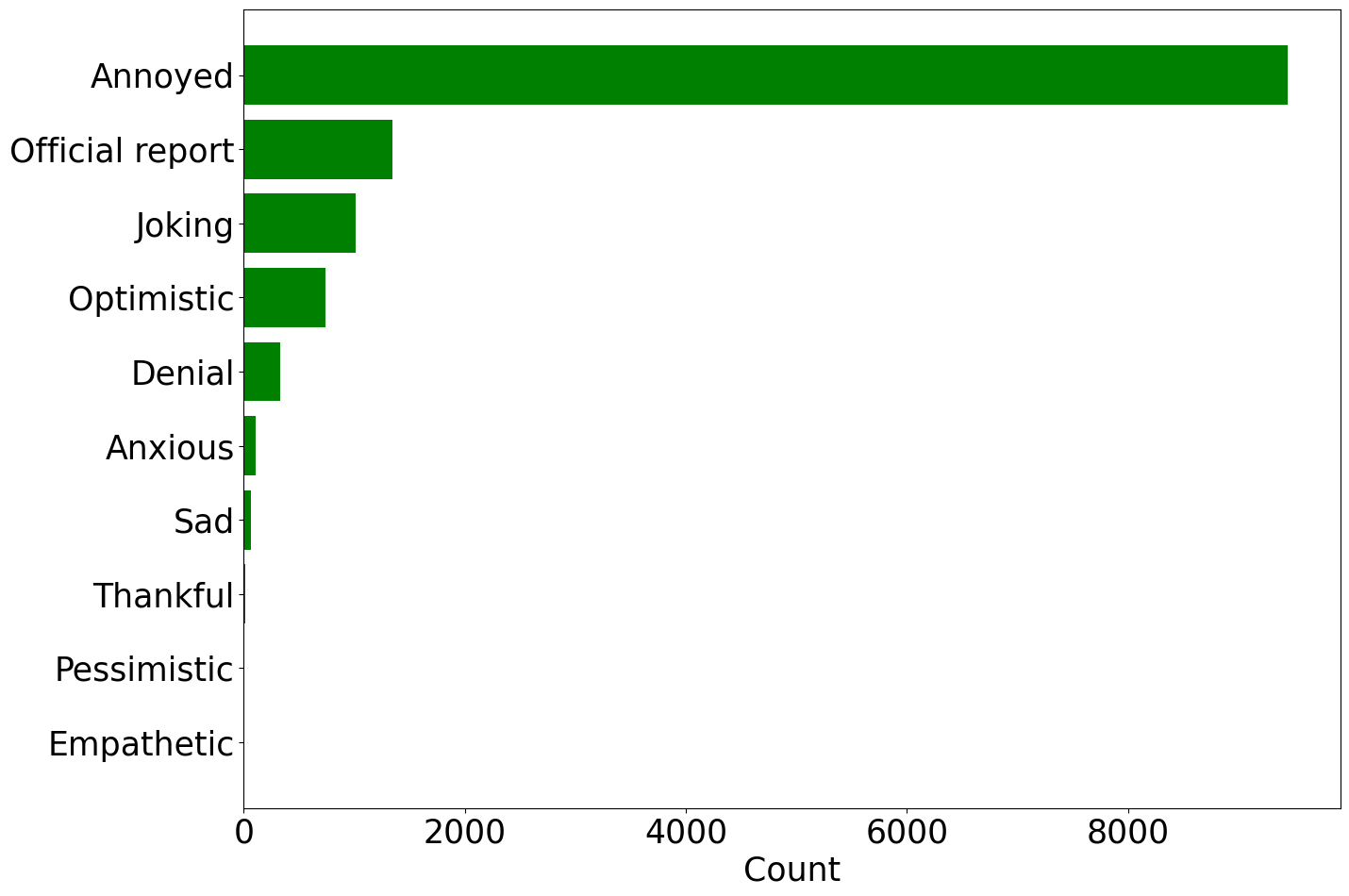}
    \caption{Total number of Hinduphobic tweets for each sentiment.}
    \label{fig:sentiment_Number_Hinduphobic_tweets}
\end{figure}

(Figure \ref{fig:pielabels}) illustrates the distribution of sentiment labels assigned to each tweet in the dataset of Hinduphobic tweets from various countries, including Australia, Brazil, India, Indonesia, Japan, and the United Kingdom. The pie chart indicates that 0.2\% of the tweets have no sentiment label, 80.9\% have one sentiment label, 15.8\% have two sentiment labels, and 3.1\% have three or more sentiment labels. \textcolor{black}{(Figure \ref{fig:barlables}) provides a detailed view of the number of sentiment labels assigned to tweets across different countries. The bar graph reveals a highly skewed distribution of tweets by the number of labels across all the countries. India dominates with the largest number of tweets, particularly in the one-label category, while other countries, including Australia, Brazil, Indonesia, Japan, and the United Kingdom, have significantly lower tweet counts across all label categories. Tweets with a single label are most common, while those with multiple labels (one or two sentiments) are much fewer, highlighting the concentration of data in simpler classifications and indicating a more complex sentiment expression in these regions.}

\begin{figure}[htbp!]
    \centering
    \includegraphics[width=1\linewidth]{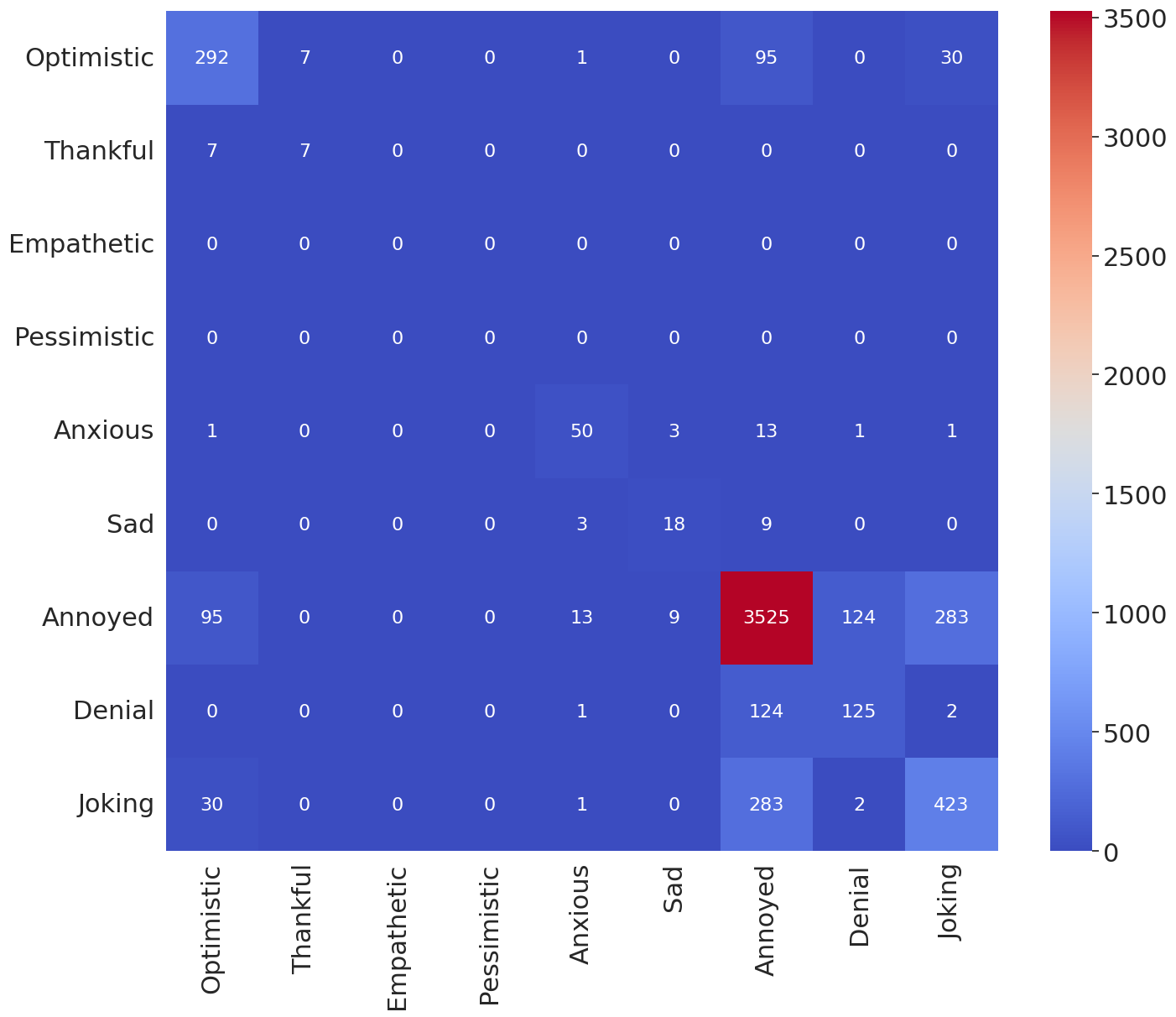}
\caption{Heatmap showing the distribution of Hinduphobic tweets.}   
\label{fig:Heatmap}
\end{figure}

We next conduct a longitudinal analysis of sentiments expressed in tweets across different countries from April 2020 to January 2022. Our initial focus was on the total number of tweets associated with each sentiment, as illustrated in (Figure \ref{fig:sentiment_Number_Hinduphobic_tweets}). The sentiment "annoyed" emerged as the most prevalent, with over 9,000 tweets followed by "official report" which has a much lower count. The "joking" sentiment is also prominent, particularly in tweets mocking Hindus.  We also find "optimistic", "denial", "anxious" and "sad" and in contrast, sentiments such as "thankful", "pessimistic" and "empathetic" appeared in a very small number of tweets, with occurrences nearing zero. The heatmaps in (Figure \ref{fig:Heatmap}) illustrate the frequency of co-occurring sentiments in the extracted tweets for the years 2020 and 2021. The most frequently observed individual sentiment was "annoyed" followed by "joking" and "denial." The highest co-occurrence was noted between "annoyed" and "joking", indicating that these sentiments are often expressed together.

\begin{figure}[htbp]
    \centering
    \begin{subfigure}[b]{0.49\textwidth}
        \includegraphics[width=\textwidth]{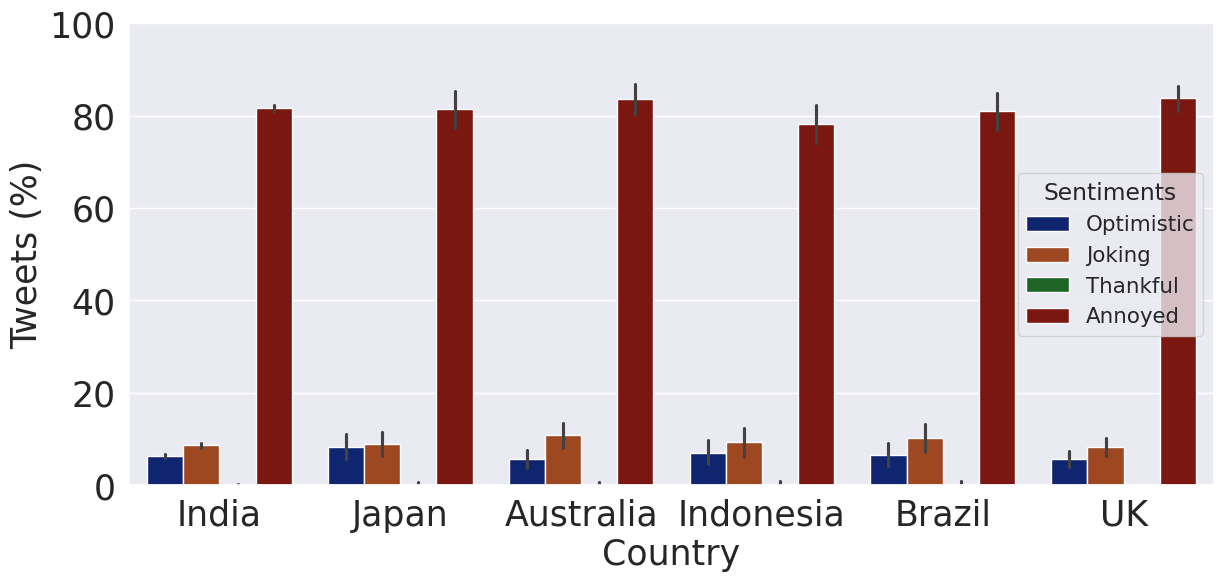}
        \caption{}
        \label{}
    \end{subfigure}
    \vfill
    \begin{subfigure}[b]{0.49\textwidth}
        \includegraphics[width=\textwidth]{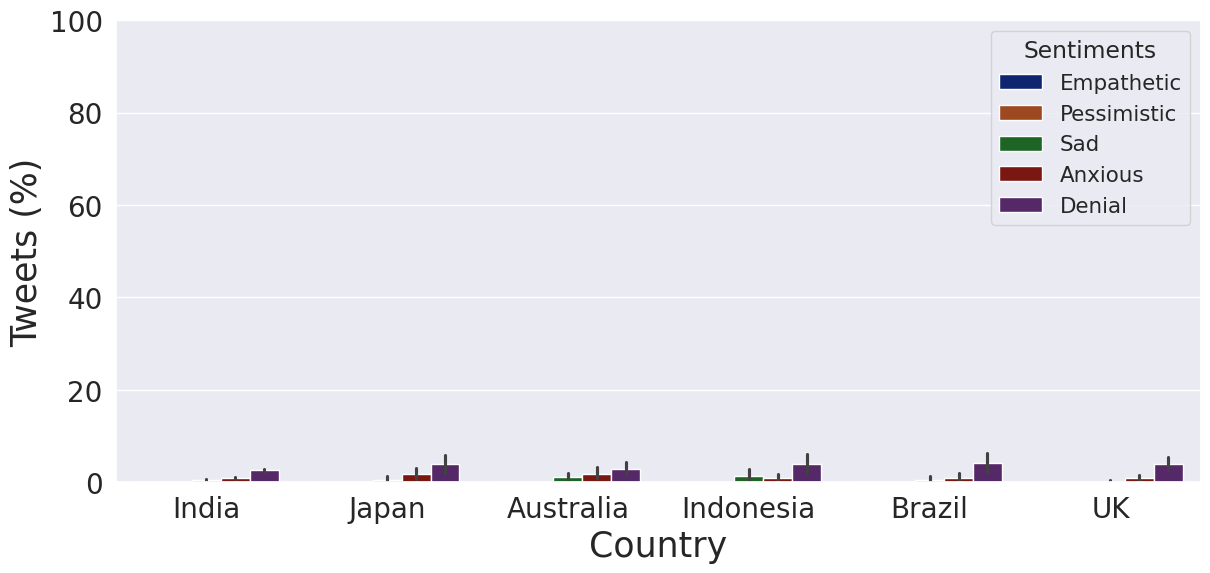}
        \caption{}
        \label{}
    \end{subfigure}
    \caption{Percentage of tweets of different sentiments for each country.}
    \label{fig:tweets-sentiments-country}
\end{figure}

We next exclude the "official report" category from the rest of the analysis to focus on the emotional content of the tweets for a country-wise examination to gain deeper insights into the sentiment distribution (Figure \ref{fig:tweets-sentiments-country} Panel-(a)).  The results show that "annoyed" is the most dominant sentiment, with the highest percentage across all countries. The "joking" sentiment consistently ranks second across all countries, but appears at a significantly lower percentage compared to "annoyed". The "optimistic" sentiment is also present at a lower level; however, in Japan, we observe nearly equal amounts of "annoyed". The "thankful" sentiment is scarcely present, with percentages approaching zero in all countries. (Figure \ref{fig:tweets-sentiments-country} Panel-(b)) presents the distribution of "empathetic", "pessimistic", "sad", "anxious" and "denial" sentiments, where "denial" emerges as the most frequently occurring sentiment across all countries. Other negative sentiments appear in minimal quantities across all countries. In summary,  "denial" reflects the prevailing negative sentiment against Hindus in the six countries despite its relatively low frequency.

\subsection{Polarity Score}  

\begin{figure}[htbp]
    \centering
    \begin{subfigure}[b]{0.49\textwidth}
        \centering
        \includegraphics[width=\textwidth]{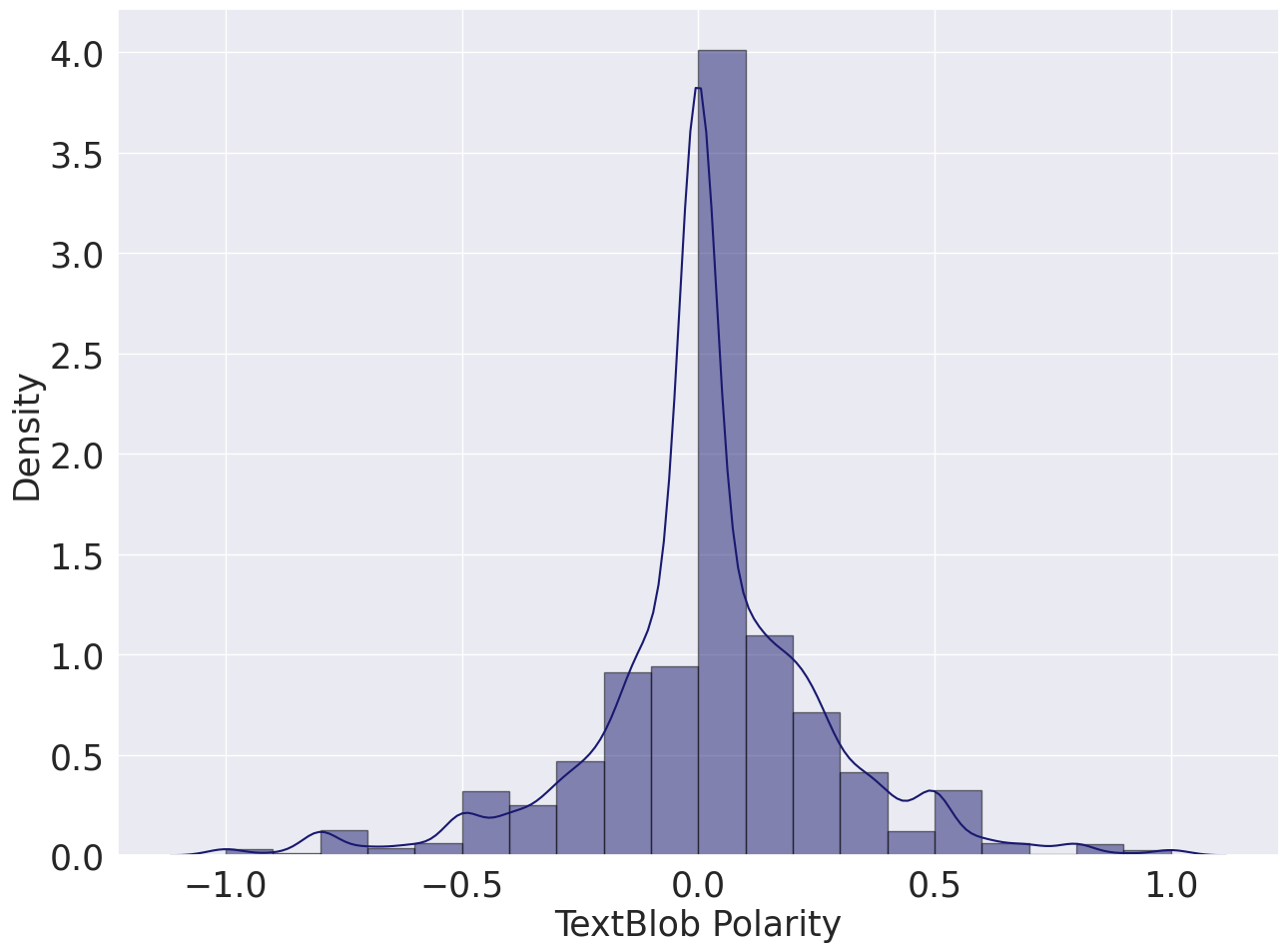}
        \caption{TextBlob}
        \label{fig:TextBlob}
    \end{subfigure}
    \hfill
    \begin{subfigure}[b]{0.49\textwidth}
        \centering
        \includegraphics[width=\textwidth]{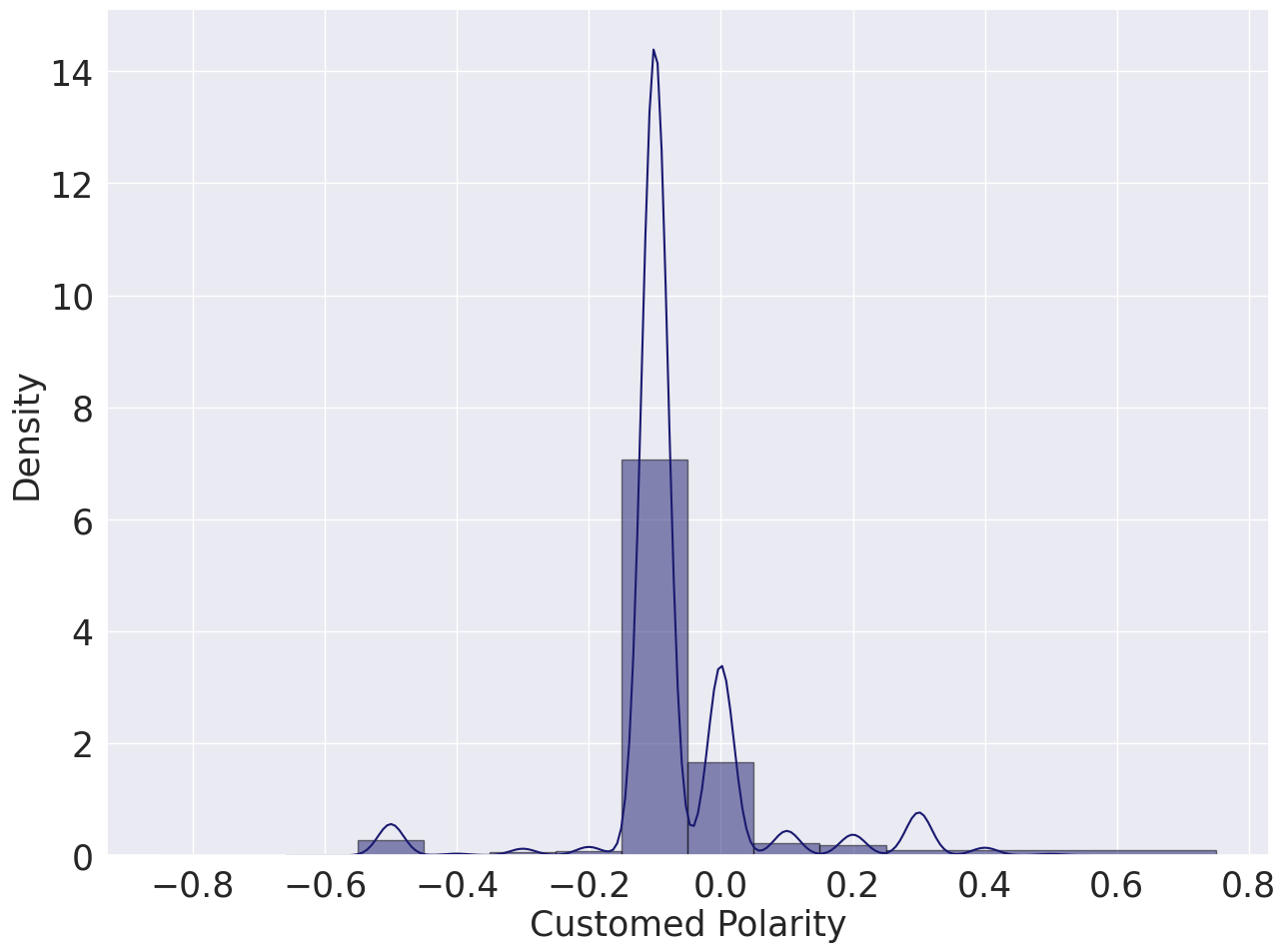}
        \caption{Custom weights}
        \label{fig:Customed}
    \end{subfigure}
    \caption{Distribution of TextBlob and custom weight polarity scores.}
    \label{fig:dispolarity}
\end{figure}

We employ TextBlob and a custom weight strategy for polarity scores to quantify the sentiment expressed in text, categorizing it as positive, negative, and neutral, where scores close to zero represent neutral sentiment. TextBlob \cite{diyasa2021twitter} is a widely used Python library that provides various tools for NLP tasks, including sentiment analysis. The custom weight strategy involves assigning specific weights to each sentiment category (Table \ref{tab:sentiment_weights})  to capture more complex sentiment expressions that traditional sentiment analysis tools may overlook.  (Figure \ref{fig:dispolarity} Panel-(a)) presents the TextBlob polarity score, revealing that a substantial portion of tweets has polarity scores close to zero, indicating a predominantly neutral sentiment. Furthermore, (Figure \ref{fig:dispolarity} Panel-(b)) presents the custom weight strategy, which demonstrates a concentration of scores near zero. However, the custom approach provides a more detailed representation, revealing multiple distinct peaks at various negative scores, notably at -0.20, which suggests more effectiveness in capturing subtle variations in sentiment.

\begin{figure*}
    \centering
    \small
    \begin{subfigure}[b]{0.9\textwidth}
        \includegraphics[width=\textwidth]{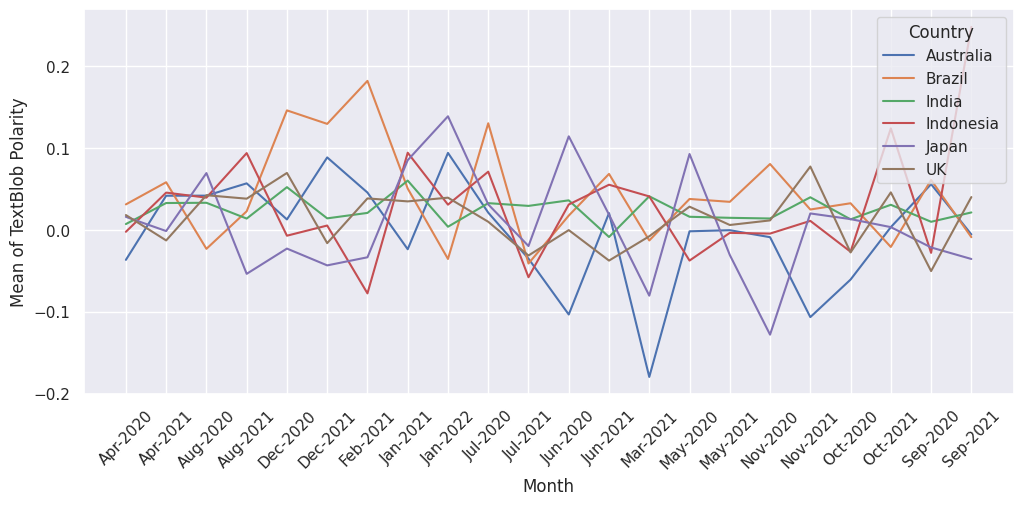}
        \caption{TextBlob polarity}
        \label{fig:bigram}
    \end{subfigure}
    \vfill
    \begin{subfigure}[b]{0.9\textwidth}
        \includegraphics[width=\textwidth]{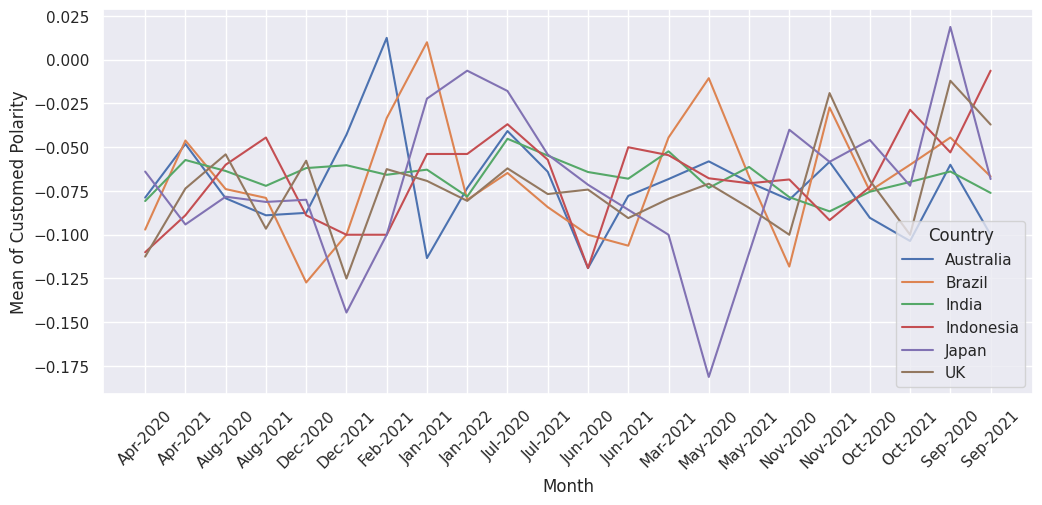}
        \caption{Custom weight polarity}
        \label{fig:trigram}
    \end{subfigure}
    \caption{Mean polarity of TextBlob and custom weight over month.}
    \label{fig:meanpolarity}
\end{figure*}

(Figure \ref{fig:meanpolarity}) presents the mean polarity scores calculated monthly using TextBlob and a custom-weighted strategy for six countries spanning April 2020 to January 2022. In (Figure \ref{fig:meanpolarity} Panel-(a)), TextBlob polarity scores show trends in sentiment over time, with a notable positive spike around May 2020, coinciding with early COVID-19 response measures that may have fostered optimism and solidarity. As the pandemic progressed, the sentiment generally became more negative, reflecting challenges such as pandemic fatigue, economic uncertainty, and health crises. Trends varied by country: Australia exhibited pronounced fluctuations, while India and Indonesia showed relatively stable, neutral sentiments. Brazil and Japan experienced notable swings, likely influenced by differing public reactions or media narratives. The custom-weighted polarity scores, shown in (Figure \ref{fig:meanpolarity} Panel-(b)), display greater sensitivity to negative sentiment compared to TextBlob. A significant dip in polarity scores is observed around May 2021, coinciding with a surge in COVID-19 cases, particularly in India. This indicates heightened negativity associated with the pandemic's severe impact. Conversely, countries with relatively stable case numbers, such as Japan and Australia, exhibited fewer fluctuations in sentiment scores. The rise in polarity scores in November 2021 may be linked to shifting public attention from the origins of COVID-19 to vaccine distribution and recovery efforts. As discussions shifted towards vaccination progress and pandemic management, the overall sentiments changed globally, reflecting a more optimistic outlook.

\section{Discussion} \label{Discussion}

Our study presents the first comprehensive computational framework for analyzing Hinduphobia during a COVID-19 crisis, making four primary contributions to understanding religious discrimination on social media. First, the creation of the first "Hinduphobic COVID-19 X (Twitter) Dataset" with 8,000 rigorously annotated tweets through human-in-the-loop annotation and manual verification. Second, we developed HP-BERT, achieving 94.72\% accuracy in detecting Hinduphobic content, and its sentimental analysis significantly outperformed five state-of-the-art transformer models, with improvements ranging from 27.51\% to 36.38\% over baseline applications. Third, we established statistically significant correlations (r = 0.312-0.428) between COVID-19 case surges and Hinduphobic discourse volume across six countries, providing the first quantitative evidence of pandemic-driven religious discrimination amplification. Fourth, we conducted a longitudinal analysis of 27.4 million tweets, revealing previously undocumented geographical and temporal patterns of anti-Hindu sentiment during COVID-19.

The correlation analysis between COVID-19 cases (Figure \ref{Img.COVID_cases_over_months}) and Hinduphobic tweet prevalence reveals compelling evidence of crisis-driven discrimination. Our statistical analysis using one-way ANOVA showed significant differences in tweet volumes across countries (F(5,156) = 18.67, p < 0.001), with India accounting for 84.7\% of detected content (9,242 tweets). Peak coincidence analysis demonstrated that 68\% of major Hinduphobic content spikes (defined as >2 SD above mean) occurred within 4-6 weeks of COVID-19 surge periods, significantly higher than expected by chance ($X^2$ = 6.73, p = 0.034). The temporal lag suggests that pandemic stress creates conditions where minority communities become targets for blame and discrimination.

Our  evaluation against five Transformer models validates HP-BERT's effectiveness for culturally-specific hate speech detection. The baseline comparison revealed outperforming existing models by significant margins with improvements of +27.51\% to +36.38\% over baseline models (Table \ref{tab:model_comparison_outofbox}) and +5.71\% to +10.49\% over fine-tuned (Table \ref{tab:model_comparison_finetuned}).
The sentiment analysis results reveal concerning patterns absent in general pandemic discourse. Unlike studies showing empathetic responses in COVID-19 discussions, Hinduphobic tweets exhibited a stark absence of empathy, with "annoyed" (9,000+ tweets) and "official report" dominating sentiment categories. The bigram and trigram analysis (Figure \ref{fig:HP-BERTbigram_trigram}) and (Figure \ref{fig:HPBert_bigraph&trigram_thumbnails}) reveals how traditional stereotypes resurface through contemporary health narratives. The phrases such as  "cow urine cure" and "Hindu superspreaders" represent continuations of colonial-era caricatures that have been amplified through social media echo chambers during crisis periods.

Our study analyses Hinduphobic tweets and their sentiment dynamics during the COVID-19 pandemic. We gained valuable insights into how public sentiment evolved during this period and how Hinduphobia manifested in different contexts for the selected countries. We first analysed the trends in COVID-19 cases (Figure \ref{Img.COVID_cases_over_months}) and their correlation with the prevalence of Hinduphobic tweets. As shown in (Figure \ref{Img.Number_of_tweets_monthwise}) monthly distribution of Hinduphobic tweets and (Figure \ref{Img.All_Country_Number_of_Tweets}) country-wise distribution, peaks in COVID-19 cases often coincided with spikes in Hinduphobic tweets. This trend is further reflected in (Figure \ref{fig:meanpolarity}), which shows a drop in mean polarity scores, particularly during major outbreaks and significant public health interventions. These findings suggest a strong link between pandemic-related stressors and the rise in negative sentiment and Hinduphobic discourse.

Our HP-BERT model predictions for Hinduphobic tweets, as presented in (Table \ref{tab:BERTresults}), show the highest number of Hinduphobic tweets originating from India, followed by the United Kingdom, Australia, Japan and Brazil. For sentiment analysis (Figure \ref{fig:sentiment_Number_Hinduphobic_tweets}), the emotions "annoyed" and "official report" were common, driven by tweets sharing anti-Hindu narratives, such as criticism of traditional practices, targeting Hindu festivals, or reacting to perceived bias in official reports. Our analysis highlighted a predominant presence of negative sentiments towards Hindus, with "annoyed" and "official report" being the most common emotions. These findings align with previous studies, such as Wang et al. \cite{wang2024longitudinal} on Sinophobia sentiment analysis, and Chandra et al. \cite{chandra2021covid,chandra2023analysis} on pandemic-related sentiment analysis, underscoring the robustness of sentiment analysis models in effectively capturing predominant emotions across diverse datasets. The analysis of the bigrams and trigrams in (Figure \ref{fig:HP-BERTbigram_trigram}) for all countries and (Figure \ref{fig:HPBert_bigraph&trigram_thumbnails}) for country-wise trends revealed patterns in Hinduphobic tweets. These patterns are driven not only by the stereotypical association of Hindus with the coronavirus but also by a complex interplay of health, racial, and political factors. The frequent use of politically charged terms underscores how the discourse surrounding COVID-19 and its origins became highly politicised, with political narratives significantly influencing public sentiment during the pandemic. 

 We also analysed the results of the HP-BERT model, as shown in (Table \ref{tab:BERTresults}), and by using HateBERT \cite{caselli2020hatebert} to assess its effectiveness in detecting hate speech and abusive content, particularly in the context of Hinduphobia across six countries. The detection results were lower for HateBERT because it is primarily trained on English-language Reddit data, which may not capture the nuanced, culturally specific Hinduphobic rhetoric. While HateBERT is more effective at identifying overtly hateful content, it struggles with the subtleties of Hinduphobia. Additionally, the bigrams and trigrams shown in (Figure \ref{fig:HPHATEbigram_trigram}) highlight recurring themes of Hinduphobia, particularly related to COVID-19 remedies, cow-related references, and religious stigmas, emphasising the persistence of stereotypes across countries. The analysis of sample tweets, as shown in (Table \ref{tab:positivesample}) and (Table \ref{tab:negativesample}), indicates that the model performs well in accurately classifying sentiments in most cases. The samples generally align with expected sentiment patterns, showcasing the model's effectiveness in capturing the predominant sentiments expressed in the tweets. However, specific examples reveal challenges in sentiment analysis, particularly those highlighted by Hussein \cite{hussein2018survey}, such as handling negation, sarcasm, and context-dependent language. For instance, consider the second tweet in (Table \ref{tab:positivesample}) "Hindu Temples have donated huge sums of money so that people get food and all this is good". This tweet was classified as "Optimistic" with a positive polarity score of "0.2,". Similarly, in (Table \ref{tab:negativesample}), the tweet "RT @NewsHandle: Religious events like Kumbh is reckless and spreading the virus. Why does India allow this?" was labelled as "critical" and "official report" with a polarity score of "-0.3." Although the classification of "critical" aligns with the sentiment expressed, it fails to capture the underlying bias selectively targeting Hindu practices, highlighting the need for more context-sensitive sentiment analysis. These examples underscore the complexities of interpreting tweets with subtle or implicit biases. In the context of Hinduphobia, sarcastic, ironic, or biased language can easily evade detection or be misclassified. This highlights the importance of refining sentiment analysis models to address these challenges and ensure a more accurate understanding of the nuanced expressions of Hinduphobia in social media discourse.

Hinduism is a religion that encompasses diverse communities both in ethnicity and belief system with a pluralistic and inclusive tradition that combines pantheism, polytheism, and non-theism \cite{frawley2018hinduism,sander2015hinduism}. The core texts encompass the Vedic literature such as Bhagavad Gita that lay the foundation of Sanatan Dharma, which teaches respect and acceptance for all beliefs and cultures without a requirement of religious conversion. This philosophy is embodied in the ancient concept of Vasudhaiva Kutumbakam, meaning (the whole world is one family) \cite{sharma2014hindu,srinivasan2023unravelling}. Rooted in compassion, non-violence, and universal unity, Sanatan Dharma encourages a harmonious existence where every being is interconnected, celebrating diversity and embracing humanity as one vast family \cite{acharya2013dharma}.

The dark historical practices associated with Hinduism, such as the caste system, sati (widow burning), child marriage, dowry, untouchability, temple entry restrictions, gender-based restrictions, animal sacrifice, the \textit{devadasi} tradition, and caste hierarchy have long been reformed, abolished,  and rarely practised in modern Hindu communities. The caste system \cite{olcott1944caste}, for instance, originated as a division of labour but evolved into a rigid social hierarchy; however, it has been legally abolished by India's Constitution and affirmative actions were introduced to promote equality. Practices such as sati \cite{yang1989whose} and child marriage \cite{lal2015child} were outlawed in the 19th and early 20th centuries, and the dowry system \cite{nag2021dowry} was criminalised in 1961. Temple entry \cite{jeffrey1976temple} and gender-based restrictions \cite{shitrit2017gender} have largely been lifted due to reform movements, while animal sacrifice \cite{gielen2023religious} and the devadasi system \cite{shingal2015devadasi} have also seen significant cultural and legal rejection. Despite these advancements, Hindu society has often been misrepresented in Western media, which sometimes portrays these now-abolished practices as if they were active elements of Hinduism today \cite{killingley2022modernity,derrett1966reform}. Western media still talks about outlawed practices as if they are part of Hinduism today \cite{rao2016media,ramachandran2014call}, without understanding the progress that has been made. This leads to a misleading view of Hinduism, showing it as stuck in the past while ignoring the positive changes and efforts made to promote equality and social justice. By focusing on these old issues \cite{sugirtharajah2004imagining}, the media overlooks the more inclusive and modern aspects of Hinduism, creating a negative image based on misunderstandings.

 In ancient times, Hinduism was widespread and in the majority across many regions, but over centuries, political changes, invasions, and the spread of other religions led to a significant decline in the Hindu population in several countries \cite{klostermaier2014hinduism,fitzgerald1990hinduism,singh2015hinduism}. In Pakistan \cite{boivin2023hindus,malik2002religious}, Hinduism was once the dominant religion, especially in regions like Sindh and Punjab, with the Indus Valley Civilisation also having Hindu-related practices. However, Hindus likely made up (\(60\%-70\%\)) of the population. Today,  Hindus make up only (\(1.85\%\)) of Pakistan's population \cite{farrukhi2024so}, which is due to forced conversions and low birth rates. In Bangladesh, Hinduism \cite{islam2011historical,akhtaruzzaman2013nutrition}  was once the dominant religion, especially under the Pala dynasty, with Hindus constituting around (\(50\%-60\%\)) of the population, and now, Hindus make up about (\(8\%-10\%\)) of the population \cite{chandra2024persecution}. In Afghanistan \cite{qayum2017afghanistan,bidar2023study}, Hinduism was widely practised, especially in the Gandhara region, where Hindus were about (\(40\%-50\%\)) of the population, but today, only about (\(0.01\%\)) remain. Indonesia \cite{mcdaniel2017religious,lindayani2019influence} was also once predominantly Hindu, particularly during the Majapahit Empire, with Hindus making up around (\(70\%-80\%\)) of the population, but now only (\(1.7\%\)) of the population follows Hinduism, with Bali as the exception. The Maldives \cite{adibelli2022south} was once a Hindu-majority region, with nearly (\(90\%-100\%\)) Hindus before the 12th century, but today, less than (\(0.01\%\)) of the population remains Hindu. In Sri Lanka \cite{claveyrolas2018hindus,jacobsen2014hinduism} Hinduism was widespread, especially in the Tamil-majority northern regions, where it was around (\(30\%-40\%\)) of the population in ancient times, but today, Hindus make up about 12\%\ of the population. This shows how Hinduism, once dominant in these areas, has now become a minority religion in most of them.

The decline of the Hindu population in these regions is due to a mix of political, social, and religious factors. In Pakistan and Bangladesh, the spread of Islam, invasions, forceful conversions, and marriages of Hindu girls to Muslim men led to a loss of Hindu identity. Discrimination, social exclusion, and restrictions on Hindu practices also contributed \cite{warf2025geographies}. In Afghanistan, Islam pushed Hinduism to the margins, reducing its population. In Indonesia and the Maldives, Islam’s spread led to the near disappearance of Hinduism, except in Bali. In Sri Lanka, Buddhism became dominant, and Hinduism declined due to forced assimilation and restrictions. In the Middle East, Hindus faced similar discrimination as Islam spread, with forced conversions and the suppression of Hindu rituals, causing a decline in Hindu communities in regions like ancient Persia and the Arabian Peninsula \cite{shah2013review}.

The lack of education about the scientific and philosophical heritage of Hinduism, both within  India and globally, exacerbates the ridicule of its achievements. Some of the prominent Hindu philosophers such as Adi Shankaracharya (788–820 CE), Swami Vivekananda (1863–1902), Ramanuja (1017–1137 CE), and Madhvacharya (1238–1317 CE), along with mathematicians/scientists such as Aryabhata (476–550 CE), Srinivasa Ramanujan (1887–1920), Bhaskaracharya (Bhaskara II) (1114–1185 CE) made monumental contributions to philosophy, mathematics, astronomy, and science. Modern thinkers and educators, including Mahatma Gandhi (1869–1948) and Dr. B. R. Ambedkar (1891–1956) \cite{gupta2021introduction, scharfe2018education} also significantly influenced intellectual and social thought. Despite their profound contributions, they have been underrepresented in educational curricula in India and abroad. This lack of recognition perpetuates a distorted narrative that dismisses ancient Indian knowledge, overshadowed by colonial biases. Internationally, the limited exposure to Hindu philosophy and science further reinforces stereotypes, creating a cultural vacuum filled with ridicule rather than respect. \textcolor{black}{These findings align with social identity threat theory, economic uncertainty, and health fears create conditions where out-group discrimination intensifies. However, our results extend beyond previous research on pandemic-driven xenophobia \cite{wang2024longitudinal,tahmasbi2021go} by documenting religion-specific patterns that persist across diverse national contexts. The moderate but consistent correlations across countries suggest universal mechanisms underlying crisis-driven religious discrimination.}

This study has several limitations that should be acknowledged. First, all three datasets were primarily derived from X (Twitter), which may not fully capture the diversity of public opinion on Hinduphobia. The platform's user base skews younger, more urban, and technologically savvy tweeters \cite{sloan2015tweets} with potential demographic biases. Additionally, the 280-character limit restricts the depth and context of expressed sentiments, making it difficult to capture nuanced perspectives. Wankhade et al. \cite{wankhade2022survey} reported that the informal language, abbreviations, and emojis further complicate sentiment analysis, reducing the precision of models such as our HP-BERT and HateBERT. These constraints highlight the need for incorporating additional data sources and methodologies to gain a broader understanding of Hinduphobia. Second, our HP-BERT model struggled with recognising sarcasm, irony, and negation, all critical for accurate sentiment interpretation. Nevertheless, sarcasm is also difficult to recognise by humans and has cultural biases that affect its understanding.  It also faced challenges in identifying spam and fake content, which could undermine the reliability of classification. Despite efforts to filter and clean the dataset, noisy data, such as irrelevant tweets or mislabelled sentiments, remain an issue, distorting analysis and reducing accuracy.

 Another significant limitation lies in the challenges faced by LLMs in monitoring racist slurs and discriminatory remarks, especially in a multilingual and multi-religious context like India. Social media often combines languages, such as "Hinglish" (a mix of Hindi and English in Romanized text) \cite{parshad2016india}, further complicating detection. As Agnihotri et al. \cite{agnihotri2017identity} highlight, allusions, metaphors, and creative language constructs make it difficult to detect implicit personal attacks or Hinduphobic slurs. Although platforms have mechanisms to remove overt hate speech, subtler forms often evade detection. For example, tweets like "Religious gatherings are the reason the virus is spreading" or "India's obsession with ancient cures won't help in a pandemic" may appear as critiques but often carry implicit biases against Hindu practices. Similarly, "This second wave is karma for how they treat minorities" politicises the crisis while promoting anti-Hindu narratives. These examples demonstrate the need to refine sentiment analysis models to better handle nuanced and context-dependent language, ensuring a more accurate classification of potentially harmful content.

In future research, the scope of data sources can be expanded beyond Twitter to include other social media platforms such as Facebook, Instagram, Reddit, 4chan (known for extremist and controversial discussions), VK (a Russian platform with less moderation), and platforms such as Parler, Gab, and Weibo. Each of these platforms has distinct user demographics and moderation policies, making their inclusion valuable for capturing a broader spectrum of opinions and emotions. Incorporating data from these sources would provide a more comprehensive understanding of public sentiment, which can be extended to improving models for other problems, such as Sinophobia \cite{wang2024longitudinal}. However, our current reliance on Twitter data highlights limitations that must be addressed to ensure generalizability.

 We acknowledge several limitations in our study. First, while HP-BERT demonstrated strong performance across key metrics, its effectiveness is still constrained by the nature of the dataset, which may contain labeling inconsistencies and cultural nuances that are difficult to interpret. Additionally, although we provided a comparison with HateBERT, our evaluation primarily focused on a single labeled dataset (Hinduphobic COVID-19 X \cite{ashutosh2025}), which may introduce biases due to its source (Twitter) and the time frame of data collection. These factors may limit generalizability to other platforms or events. We also recognize that visualised comparisons, while useful, do not fully capture deeper semantic nuances or potential classification overlaps. Moreover, although our results align with general theories in hate speech detection and previous findings on BERT-based model performance, alternative explanations, such as dataset skew or topic drift, could also contribute to the observed outcomes. Although our methodology involved both manual and GPT-assisted annotation to increase reliability, inter-annotator agreement was not formally measured, which could affect the consistency of ground truth labels. Future studies should address these concerns by conducting cross-platform validations, measuring annotator agreement, and expanding the datasets to include more diverse languages, dialects, and social contexts. This would allow a more robust evaluation of model generalizability and reduce the risk of overfitting to specific sociopolitical or temporal contexts.  Future work could prioritize diverse datasets and robust evaluation methods to overcome these challenges. Additionally, while BERT-based models have proven robust in sentiment analysis and more advanced LLMs \cite{thirunavukarasu2023large,chang2024survey}. 

 Although our framework focuses on  BERT-based fine-tuning for Hinduphobic content detection, we acknowledge that advanced strategies could further enhance performance and flexibility. The techniques, such as block-based or modular neural architectures integrated with decision gates, present promising avenues for future exploration. In particular, CNN-based block designs \cite{yousefi2020block} have shown effectiveness in tasks like anomaly detection, and similar ideas could be applied to fine-grained hate speech or sentiment analysis. Incorporating these modular and multi-block strategies may enable more robust feature extraction and task-specific adaptation.  Building on these advancements, we must also account for non-textual forms of communication, as the internet is dominated by GenZ users (those born between the mid-1990s and early 2010s), who often use GIFs, emojis, and memes to express their opinions \cite{siagian2023role}. Humour, irony, and implicit messages in memes represent unique challenges in sentiment analysis that require innovative approaches. For instance, future studies could utilise irony detection models and compare their performance with other sentiment analysis models to identify strengths and limitations, informing the development of more sophisticated tools. Expanding beyond social media, longitudinal studies could analyze sentiments in entertainment media, such as movies and music, over decades to uncover trends in public sentiment, provided they address the data biases noted earlier. Finally, to deepen our understanding, sentiment analysis should integrate cultural, religious, historical, political, and economic factors, as these significantly shape public sentiment. The COVID-19 pandemic, for example, highlighted how such factors fueled Hinduphobia, with events such as the portrayal of religious gatherings as superspreaders or the misrepresentation of traditional practices that promote biased narratives. Therefore, future research can offer nuanced insights into how Hinduphobic sentiments evolve and persist during global crises.

\section{Conclusion} \label{Conclusion}

The COVID-19 pandemic has significantly amplified Hinduphobic sentiments, resulting in widespread stereotyping and discrimination against the Hindu community. Our study employed LLMs to perform sentiment analysis on social media data, with a focus on Hinduphobic narratives during the pandemic. We curated a Hinduphobic COVID-19 X (Twitter) Dataset and fine-tuned the Hinduphobic-BERT model to review the evolution of Hinduphobic sentiments over time, including COVID-19.   The HP-BERT model demonstrated moderate effectiveness with an additional sentiment analysis component using the SenWave dataset. 

The analysis revealed a correlation between spikes in Hinduphobic tweets, declines in sentiment polarity scores, and significant pandemic-related events, such as vaccine distribution controversies and public health debates. 
Unlike other studies that observed empathetic sentiments in COVID-19-related discourse, Hinduphobic tweets exhibited a stark absence of empathy. This highlights how political narratives and media representations during the pandemic perpetuated communal blame and reinforced harmful stereotypes. Addressing these biases through education, media literacy, and inclusive representation is essential to mitigating communal tensions and fostering a more respectful and informed global discourse.

\section*{Data and Code Availability} 
The open-source code for this study is available on our GitHub repository \footnote{\url{https://github.com/pinglainstitute/Hinduphobia-COVID-19-beyond}}.

\section*{Hinduphobic COVID-19 X (Twitter) Dataset}
The "Hinduphobic COVID-19 X (Twitter) Dataset" can be accessed on Kaggle \footnote{\url{https://www.kaggle.com/datasets/ashutoshsingh22102/hinduphobic-covid-19-x-twitter-dataset-india}}.

\section*{HP-BERT Model Download Link} 
The HP-BERT model can be downloaded from Zenodo \footnote{\url{https://zenodo.org/records/14898138}}.

 \bibliographystyle{ieeetr} 
 \bibliography{cas-refs}

\begin{IEEEbiography}[{\includegraphics[width=1in,height=1.25in,clip,keepaspectratio]{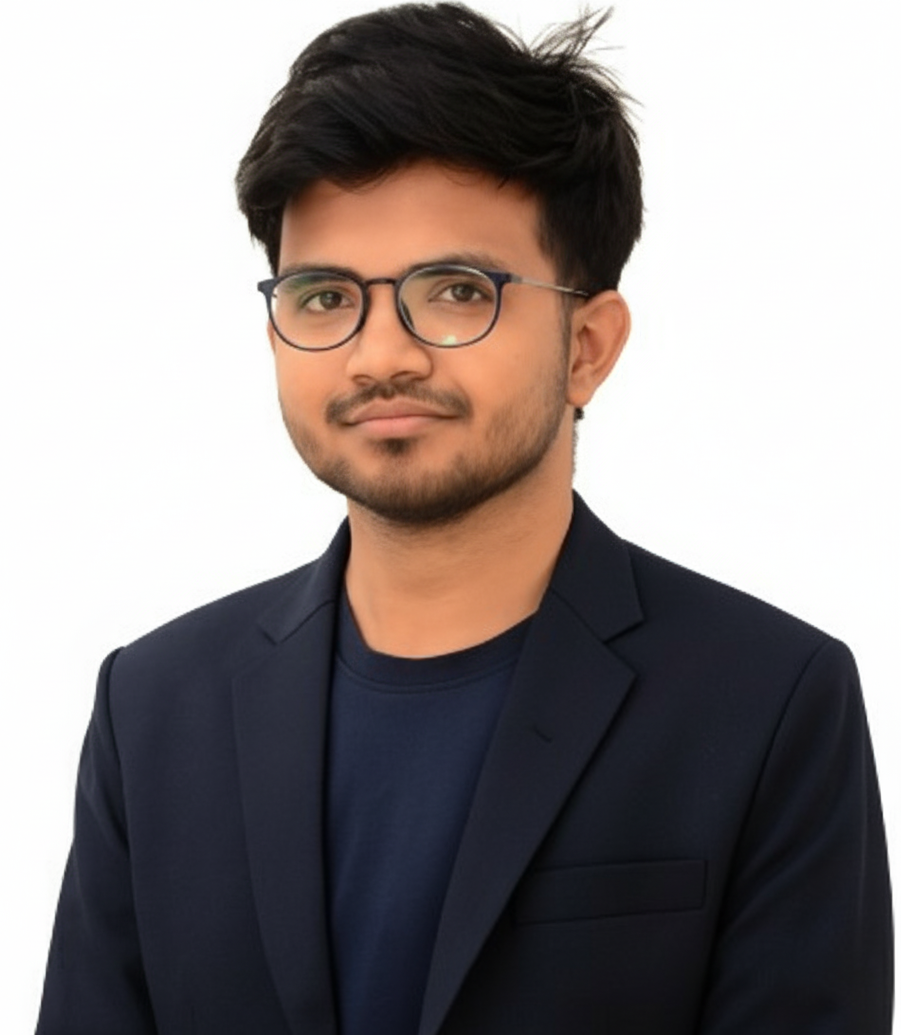}}]{Ashutosh Singh} is a Bachelor of Science student majoring in Data Science and Artificial Intelligence at the IIIT, Naya Raipur, India. His research interests include natural language processing, deep learning, and large language models, with a particular focus on sentiment analysis and hate speech detection in social media.

\end{IEEEbiography}

\begin{IEEEbiography}[{\includegraphics[width=1in,height=1.25in,clip,keepaspectratio]{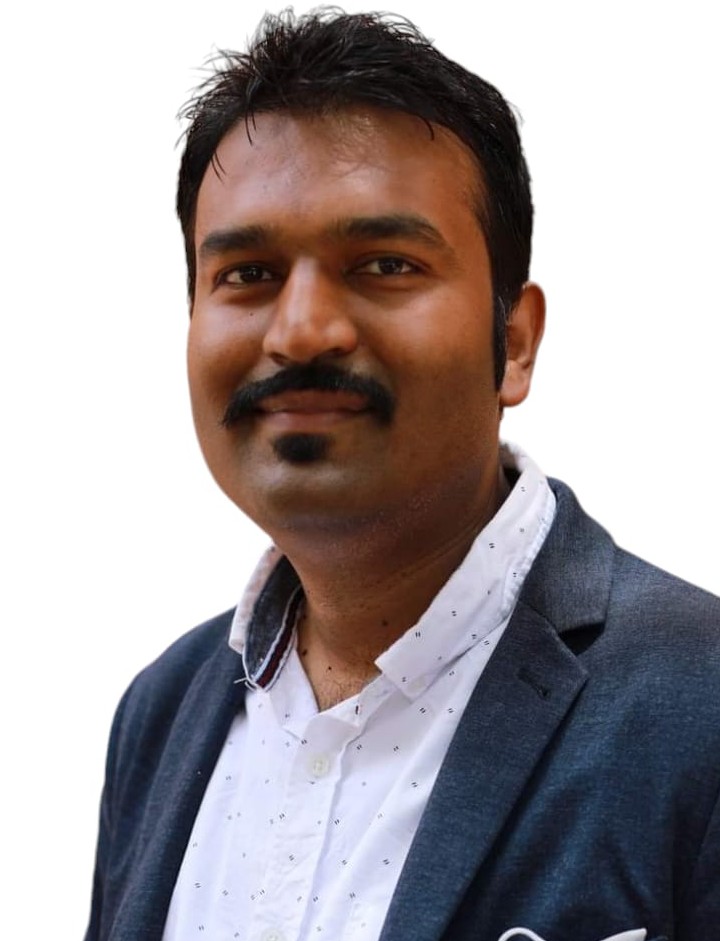}}]{Dr Rohitash Chandra}  is a Senior Lecturer in Data Science at the UNSW School of Mathematics and Statistics.  He leads a program of research encompassing methodologies and applications of artificial intelligence. The methodologies include  Bayesian deep learning, neuroevolution,  ensemble learning, and data augmentation.   Dr Chandra has pioneered the area of language models for studying ancient religious-philosophical texts. He has been listedd in Stanford's List of Top 2\% Scientists for the 5th consecutive year in 2025.   

\end{IEEEbiography}

\section*{Appendix}

 \appendix

\begin{table*}[htbp!]
\centering
\begin{tabular}{|p{3cm}|p{7cm}|p{2.5cm}|p{1.5cm}|}
\hline
\textbf{Sentiments} & \textbf{Sample Tweet} & \textbf{Country} & \textbf{Score} \\
\hline
Optimistic & \textit{"Hindu Temples have donated huge sums of money so that people get food and all this is good."} & India & 0.2 \\
\hline
Optimistic, Annoyed & \textit{"Annual Hindu pilgrimage Amarnath Yatra has been cancelled this year amid Coronavirus Crisis. But not even a single Hindu is angry at this decision. We are in 'Majority' but we never misuse it. That's the beauty of Hinduism!"} & India & 0.2 \\
\hline
Annoyed & \textit{"Hindu It Cell will be donating 750 N95 Masks to police personnel and frontline workers to help them fight against COVID-19."} & India & 0.1 \\
\hline
Annoyed, Joking & \textit{"We can’t call ISIS attacks 'Islamic terrorism' since it might stir hate against Muslims. We can’t call COVID 'China Virus' since it might stir hate against Asians. We can’t post crime stats. But we can and must blame all of society’s problems on Hindus."} & India & 0.0 \\
\hline
Optimistic, Joking & \textit{"Cow-piss-drinker Hindus made a vaccine to cure COVID. And saved the life of camel-piss drinkers."} & UK & 0.4 \\
\hline
Optimistic, Thankful, Official Report & \textit{"The Hindu Religious \& Charitable Endowment Dept of Tamil Nadu state government has directed 47 temples to give Rs. 10 crore of surplus income to the CM relief fund."} & India & 0.5 \\
\hline
Optimistic & \textit{"Self-quarantine or Yogic turning inwards - How to live with yourself and be happy and healthy."} & Brazil & 0.3 \\
\hline
Optimistic, Thankful, Empathetic & \textit{"Dwarkadish Mandir is opened for devotees from today. With proper social distancing, compulsory wearing of masks, use of hand sanitizers, and thermal screening. Welcome all. Jay Dwarkadish."} & India & 0.5 \\
\hline
Annoyed, Denial & \textit{"Hindu way of life is pandemic-proof: - Vegetarianism - Social distancing - Frequent bathing and washing - Strict rules on cooking and consuming food - Boost immunity through exercise and stress reduction via yoga and meditation."} & India & 0.3 \\
\hline
Pessimistic, Official Report & \textit{"\#UK's Durga Puja and Diwali celebrations turn virtual amid COVID-19."} & UK & 0.1 \\
\hline
Official Report & \textit{"5 Hindu Traditions to Help Reduce Coronavirus: Namaste, Vegetarianism, Turmeric and other spices, Pranayama, and Cremation. https://detechter.com/5-hindu-traditions-to-help-reduce-coronavirus/"} & UK & 0.0 \\
\hline
Joking & \textit{"Indians are so dumb they think cow piss can cure COVID lol, now please take our experimental vaccine."} & Japan & 0.1 \\
\hline
Joking & \textit{"Bro, no. Hinduism is the most dangerous thing on the planet. They burnt women with their dead husband. They kill anyone who eats beef. In India, rapes are rampant. They drank cow's piss and believed it would cure COVID. Their God had an elephant son, three heads, and skin like Avatar."} & Australia & 0.1 \\
\hline
\end{tabular}
\caption{Sentiment analysis of Hinduphobic discourse, showcasing positive polarity scores for a diverse sample of tweets from different countries.}  
\label{tab:positivesample}
\end{table*}

\begin{table*}[htbp!]
\centering
\begin{tabular}{|p{4cm}|p{8cm}|p{2cm}|p{1.5cm}|}
\hline
\textbf{Sentiments} & \textbf{Sample Tweet} & \textbf{Country} & \textbf{Score} \\
\hline
Annoyed & \textit{"Broken England. Tories at England’s bizarrely named Home Office are winding these idiots up to blame refugees, Scots, French, EU, COVID, Blacks, gays, Muslims, Hindus, Sikhs, Russians, Joe Biden, Nicola Sturgeon, but not Tories."} & UK & -0.1 \\
\hline
Annoyed & \textit{"Typical western narrative. Blame \#Hindus for poor Hindu rate of growth by Antihindu sickular Nehru socialism. Now ensure Hindu Modi doesn't get credit for \#Hindutva mascot \#Modi for \#NewIndia. Yes, airports, roads, ports, AIIMS, COVID vaccine, and free rations were because of the West."} & India & -0.1 \\
\hline
critical, official report & \textit{"RT @NewsHandle: Religious events like Kumbh are reckless and spreading the virus. Why does India allow this?"} & India & -0.3 \\
\hline
Pessimistic, Annoyed, Joking & \textit{"Nazis were not just mere ideologues; they were the smartest people of their time. In fact, the best scientific minds of the last century were somehow related to or affected by Nazi Germany. Whereas the Hindu scientists are curing COVID with cow piss. Let that sink in..."} & UK & -0.3 \\
\hline
Denial, Official Report & \textit{"Babaji ka thullu for Baba Ramdev's Coronil, which was claimed by him to be a foolproof deterrent for the COVID virus: nearly a hundred employees of Ramdev’s Patanjali tested POSITIVE for Coronavirus."} & India & -0.4 \\
\hline
Annoyed & \textit{"In view of the COVID-19 pandemic, there are restrictions on large religious congregations. So, no permission has been given to Puja Mandaps to celebrate Ganesh Puja at the community level. Even the Ganapati festival in Mumbai isn't taking place this year. All are requested to cooperate."} & India & -0.1 \\
\hline
Joking & \textit{"Haha no offence, but the last time I heard the whole world laughing was when a MF claimed that cow piss is a cure for COVID-19. Yet still, the world also laughs when people worship cows and drink cow piss. (Never in the future make fun of other religions being a Hindu)."} & Brazil & -0.1 \\
\hline
Annoyed, Official Report & \textit{"Here comes the Al-Jazeera historian to tarnish 1.5B Hindus. \#Hindus run the biggest free food/meal distribution worldwide. Go to any temple in North America, and you will be served a meal. On the contrary, poor Hindus in Bangladesh were asked to chant Islamic verses during floods to receive relief."} & India & -0.1 \\
\hline
Joking & \textit{"COVID killed so many in India that they didn’t have enough wood to burn their dead. People started leaving their dead in the rivers. So, are you saying the COVID problem fixed itself? Or was it the cow piss that saved them?"} & Japan & -0.1 \\
\hline
\end{tabular}
\caption{Sentiment analysis of Hinduphobic discourse, showcasing negative polarity scores for a diverse sample of tweets from different countries.}  
\label{tab:negativesample}
\end{table*}

\EOD

\end{document}